\documentclass[10pt]{article}
\usepackage{fullpage}
\usepackage{hyperref}
\usepackage{xcolor}
\usepackage{enumitem}
\usepackage{makecell}\usepackage{graphicx}
\usepackage[backend=biber,style=authoryear,autocite=inline,maxbibnames=10]{biblatex}
 \usepackage{algpseudocode,algorithm}
 \usepackage{amsfonts}
\usepackage{amsmath}
\usepackage{amssymb}
\usepackage{color}
\usepackage{mathtools}
\usepackage{tabularx}

\newcommand{\pair}[2]{\ensuremath{\langle #1, #2 \rangle}}
\newcommand{\env}{{\ensuremath{\mathcal{E}} }}
\newcommand{\lrn}{{\ensuremath{\mathcal{L}} }}
\newcommand{\cl}[1]{\ensuremath{\mathcal{#1}} }
\newcommand{\xv}{\mathbf{x}}
\newcommand{\wv}{\mathbf{w}}
\newcommand{\mathequiv}{{\ensuremath{\dot =}}}
\renewcommand{\Re}{\mathbb{R}}
\algrenewcommand\algorithmicuntil{\textbf{}}

\begin{document}
\title{Towards model-free RL algorithms that scale well with unstructured data}
\author{Joseph Modayil and Zaheer Abbas}
\maketitle

\begin{abstract}
Conventional reinforcement learning (RL) algorithms exhibit broad generality in their theoretical formulation and high performance on several challenging domains when combined with powerful function approximation. However, developing RL algorithms that perform well across problems with unstructured observations at scale remains challenging because most function approximation methods rely on externally provisioned knowledge about the structure of the input for good performance (e.g. convolutional networks, graph neural networks, tile-coding). A common practice in RL is to evaluate  algorithms on a single problem, or on problems with limited variation in the observation scale. RL practitioners lack a systematic way to study how well a single RL algorithm performs when instantiated across a range of problem scales, and they lack function approximation techniques that scale well with unstructured observations.

We address these limitations by providing environments and algorithms to study scaling for unstructured observation vectors and flat action spaces.  We introduce a family of combinatorial RL problems with an exponentially large state space and high-dimensional dynamics but where linear computation is sufficient to learn a (nonlinear) value function estimate for performant control. We provide an algorithm that constructs reward-relevant general value function (GVF) questions to find and exploit predictive structure directly from the experience stream.
In an empirical evaluation of the approach on synthetic problems, we observe a sample complexity that scales linearly with the observation size.  The proposed algorithm reliably outperforms a conventional deep RL algorithm on these scaling problems, and they exhibit several desirable auxiliary properties. These results suggest new algorithmic mechanisms
by which algorithms can learn at scale from unstructured data.

\end{abstract}

\section{Introduction}
An inspirational property identified by animal learning experiments is the animal's ability to make temporal associations between a seemingly arbitrary stimulus and a behaviorally relevant response within the span of a research experiment.  Though the learning of such associations was noted by early animal learning researchers~\autocite{pavlov-quote}\footnote{``The essential feature... that under different conditions these same stimuli may initiate quite different reflex reactions; and conversely the same reaction may be initiated by different stimuli.''}, the combined flexibility and speed of learning remains largely unexplained by standard machine learning theory, where standard results emphasize the exponential sample complexity of statistical estimation in high-dimensional spaces.  These empirical and formal observations are not in direct contradiction, as the learning problems addressed by animals do not span the space of problems considered by the formal theory (and vice versa).
Individual  brains also demonstrate remarkable neuroplasticity in the face of traumatic damage within the lifetime of a single animal.  Human adults show the ability to rewire brain regions that are typically associated with processing vision for audio, and the phenomenon of synesthesia demonstrates the brain can fluidly blend information from multiple modalities~\autocite{Doidge}.  Brains recovering from injuries demonstrate the capability for dynamic reorganization, by interpreting high-dimensional data from sensors that were connected by happenstance~\footnote{The  extent of plasticity in biological brains is not settled.  Some emphasize its limitations~\autocite{dehaene2020} and others its capabilities~\autocite{Eagleman2020}.  From the authors' perspective, these biological phenomena motivates the investigation into computational mechanisms that can function under similar constraints.}.  If brain regions can self organize to unstructured sensory input within the span of an individual's lifetime, then one should expect some machine learning algorithms to be capable of a similar feat.

The adaptability to unstructured inputs stands in sharp contrast to modern machine learning algorithms that rely heavily on mechanisms like convolutions or fixed positional embeddings to achieve high performance on problems with large input spaces.  This is not to say that machine learning is bereft of methods for handling unstructured inputs.  Techniques including decision-trees,  fully-connected multi-layered networks, radial basis functions, and polynomial kernels have been well-studied and commonly deployed to handle problems when linear function approximation in the inputs is insufficient~\autocite{mitchell1997machine,mackay2003information}.  However, their study is largely confined to problems with smaller input spaces or low-dimensional latent spaces, and they typically operate offline on batches of data.  There are few online function approximation techniques available to efficiently handle problems with a large unstructured space.

The danger of focusing on overly small-scale problem regimes is arguably more pronounced within reinforcement learning, as common research domains provide a mix of nearly-deterministic simulators~\autocite{machado2018revisiting}, classic problems with a small continuous state spaces~\autocite{Sutton96generalizationin}, and medium-dimensional continuous control tasks~\autocite{tassa2018deepmind}.  Both curated collections of simulation-based RL problems~\autocite{brockman2016openai,Osband2019} and randomly selected Garnet MDPs~\autocite{bhatnagar2009natural} offer more direct control over the observation size, but these domains still do not make scaling easy to study in conjunction with non-linear function approximation.

Our research question is to find problem constraints that are adequate for a single RL algorithm to perform well across problem scales, without a priori access to the input structure. 
A common misconception is that relevant input structure is always available, and thus one does not need to study problems where the structure is not available.  
Many domains produce data where the spatial arrangement of the sensors is not a good reflection of the domain dynamics, including neuron readings from brains and muscles, protein interactions, voltages in electronic systems, and internal activations of fully connected layers in artificial neural networks.  The lack of methods that reliably learn on problems with large unstructured inputs is a valid concern in such domains. 
The algorithms in this paper provide evidence that learning competent behavior from high-dimensional inputs with limited resources is possible, even when non-linear features are required and a low-dimensional latent space is not available.

 The underlying premise of this work is that some datastreams will have simple statistical regularities that expose exploitable problem structure. By uncovering computationally accessible statistical regularities in the datastream, algorithms can support the computationally scalable construction of nonlinear features that are useful for control.\footnote{A similar intuition has driven a more formal analysis~\autocite{abbe2021staircase} of the power of residual connections in neural architectures.}  In this paper, our focus is on the setting where the problem's original base features are readily identifiable for constructing relevant subproblems and their solutions.  This is a step towards the larger goal of understanding how to design RL algorithms that continue to scale well in more general settings.

\section{Problem Formulation}

This section summarizes some motivating ideas in the literature for algorithms that can scale well, provides a means to construct a family of related environments to study algorithmic scaling, and then introduces some metrics by which we can analyze how an algorithm performs across problem scales.

\subsection{Background}
To study problems of scale within reinforcement learning,  we can find guidance from planning problems. A standard planning objective is to computationally transform a given MDP with a concise formal description into a value function or policy for computationally responsive behaviour.  One form used to express the transition dynamics is the factored MDP where the transition dynamics between environment state variables are expressed by dynamic Bayes nets~\autocite{boutilier2000stochastic,koller1999computing}.  The latter paper also demonstrated that a computationally concise description of transition dynamics can still require computationally expensive form for the value function.  The paper also suggests that even when representing the exact value function is complex, an approximate value function be both simple and useful. 

Despite the limitations encountered in these works for planning (for when the transition dynamics are given), they also provide a useful starting point for examining the challenge of model-free learning when the transition dynamics are not known. In particular, we can study how well model-free RL algorithms perform at scale by restricting our attention to problems where  value function estimates can be computed efficiently using some nonlinear combination of observable features.  The learning problem can be challenging for domains that are essentially trivial for planning when the problem structure is known.  

Let us review the differences between a conventional learning problem and a planning problem.  For a planning problem, the full MDP structure is given to the optimization algorithm, including states and transition dynamics.  For a learning problem, the underlying dynamics are not given, and the environment states are accessed implicitly through observation vectors by learning algorithms using function approximation.  There are at least two forms of structure that can be directly accessed, namely the transition-dynamics in planning problems and the observation structure in learning problems.  The transition structure is exploited directly in planning problems in the papers cited above.  The observation structure is often exploited directly by learning algorithms, for example exploiting the spatial regularity of an image in a convolutional neural network.  The insight we pursue here is that a problem that is computationally trivial from a planning perspective (for example a fully factored MDP with additive rewards) will be a challenging learning problem if the observation structure is not known and non-linear feature combinations are also required.

\subsection{Constructing a scalable family of MDPs} \label{sec:problem-formalization}

We develop some desirable properties for an environment and learning algorithm pair, $\pair{\env}{\lrn}$.  We restrict our attention to continuing (non-episodic) fully-observable ergodic MDP environments, where the observation is fully determined by the state and vice versa.  Moreover, for every environment, we assume there is some function $\phi$ mapping observations to a real-valued vector. Conventionally, one learning algorithm $\lrn$ is specialized to one environment $\env$ through a choice of function approximator and hyperparameter setting selected by a researcher or machine learning practitioner.  Sometimes, the learning algorithm $\lrn$ is evaluated on a set of environments $\{\env_1, \ldots\}$, that share the same observation and action space~\autocite{machado2018revisiting}, and sometimes over varying problem interfaces~\autocite{tassa2018deepmind}.  However, there is typically little to enable the direct manipulation of the problem scale and to simultaneously support systematic comparison of results across problem scales.

We will construct a family of environments that permit systematic variation of problem scale, where the resources available to an instance of the learning algorithm can be independently varied.  In particular, we construct a family of MDPs of varying scales, 
starting with a given parent MDP $\cl{M}=(\cl{S}, \cl{A}, \cl{T}, \cl{R}, s_1)$.  We construct the family of MDPs by describing mechanisms to create new MDPs from existing MDPs, and then combine all the mechanisms.

\begin{itemize}

\item We first define a product MDP, in a similar manner to Singh and Cohn (1997)\nocite{singh1997dynamically}.  We start with $n$ individual MDPs, $\{\cl{M}^i\}_{i=1..n}$, where each MDP is given by a tuple $\cl{M}^i \mathequiv (\cl{S}^i, \cl{A}^i, \cl{T}^i, \cl{R}^i, s^i_1)$ describing the states, actions, transition dynamics, reward function, and initial state distribution.  
Restrict these to be continuing ergodic MDPs.  The product MDP considers the cross product of the component MDPs, to produce a new MDP 
$\hat{\cl{M}} = (\hat{\cl{S}}, \hat{\cl{A}}, \hat{\cl{T}},\hat{\cl{R}}, \hat{s}_1)$  where  
$\hat{\cl{S}}= \Pi_{i=1}^n \cl{S}^i$,
$\hat{\cl{A}}= \Pi_{i=1}^n \cl{A}^i$,
$\hat{\cl{R}}= \sum_{i=1}^n \cl{R}^i$,
$\hat{\cl{T}}= \Pi_{i=1}^n \cl{T}^i$, and
$\hat{s}_1= (s^1_1, \ldots, s^n_1)$, where the transition function operates on each component independently and the rewards are summed.  Planning in the product MDP is trivial as each component can be solved independently for policies or values, and the component solutions can be lifted directly  to the product.

\item Continuing to follow Singh and Cohn, we consider the problem where the set of valid actions in the product MDP is restricted to a subset of the full product.  In particular,  we restrict our attention to the setting where the action space is shared across components ($\cl{A}^i=\cl{A}^j$ for all $i,j \in\{1,..,n\}$), and the same action is broadcast to every component.  The dynamics in each component are then coupled through the action choice.  Ergodicity of the component MDPs implies the resulting MDP is also ergodic.

   \item 
    Sending the same action to many identical components will not provide a challenging environment at scale if all components are emitting rewards at a high rate.   We introduce a mechanism to control the rate at which components emit non-zero rewards.
    
     We consider MDPs where non-zero rewards are only emitted from some states. We formalize this by an indicator function $\text{hot}:\cl{S}\rightarrow \{0,1\}$ that encompasses these states ( $\cl{R}(s,a,s') \neq 0 \implies \text{hot}(s)=1$). 
    Define the set of hot states, $\cl{H} \mathequiv \{s \in \cl{S} | \text{hot}(s)=1\}$, which partitions the state space, $\cl{S} = \cl{H} \dot{\cup} \bar{\cl{H}}$. 
    
    We construct a parameterized set of MDPs that differ only in their transition probabilities into and out of these hot states,  with a fixed probability $p_{hot}$ of entering a hot state.
    \[\forall  s \in \bar{\cl{H}}\ \left( \exists s' \in \cl{H}  \ s.t.\  \cl{T}(s,a,s')>0 \implies \sum_{s''\in\cl{H}} \cl{T}(s,a,s'') = p_{hot} \right)\] 
    Such environments with state-gating of non-zero rewards can be constructed from standard MDPs.  Informally, one can augment an MDP with a binary ``hot'' feature that is used to set rewards to zero when the hot feature is off, and also define a transition probability for activating and deactivating the hot feature.  This construction provides an easy mechanism to bound the expected reward rate for policies across a set of MDPs.

\item A last augmentation is to create permuted observations and actions for an MDP.  We start with a parent observation function $\underline{\psi}:S \rightarrow \mathbb{R}^m$. 
From this function $\underline{\psi}$, we create alternate observations of the same underlying MDP from the permutation of the observations, defining $\psi_{\sigma}(s) \mathequiv \sigma(\underline{\psi}(s))$ for $\sigma \in Sym(m)$, the group of $m$ factorial permutation functions on $m$ symbols.  We can similarly construct permutations on the actions sent into an MDP, where there is a fixed permutation $\Psi_\varsigma$ for $\varsigma \in Sym(|\cl{A}|)$ applied to the actions sent into an MDP.
We define an environment, $\cl{E}$, to be the combination of an MDP $\cl{M}$ with a particular observation function, $\psi_\sigma$, and a particular action mapping, $\Psi_\varsigma$,
$\cl{E}=(\cl{M}, \psi_\sigma, \Psi_\varsigma)$.   
 \end{itemize}

We compose all of the above transformations to create combined MDP environments for this paper.  We start with a set of $n$ MDPs with a shared action space, augmenting each with a separate binary observable feature that identifies the hot states.  We take the product of the augmented MDPs and restrict the joint action space to the base action space by broadcasting.  Finally we consider the set of all environments created from permutations of a default observation function. 

Now consider how the average reward for the composite environment can be calibrated.   Note that if $p_{hot} \propto 1/n $, then the scale of the rewards from the composite MDP will be bounded if the rewards from a single MDP is bounded.  This constraint does not mean the average reward for the optimal policy or random policy is held constant (due to additional mixing time concerns from how long the system stays in $\cl{H}$), but still provides a useful bound.

The core challenge for learning a performant policy in the composite MDP is learning to associate the hot feature with the relevant observable features from each component. The underlying state space of the problem grows as $d^n$, if each component has a state space of size $d$.
Note that tabular or instance-based solution approaches require resources that grow exponentially in $n$ (to store values for every possible state).  Further note that learners based solely on linear features of the observation will not be able to associate the hot variable with features from its associated component---even if the component problems can be adequately solved with a linear learner, the combined MDP will require non-linear features for a good approximation of the optimal value function.  This construction thus provides several desirable characteristics of combinatorially large problems where computationally tractable value function estimates are possible when the learning algorithm can construct a relevant representation.
\subsection{Scaling metrics}

The remainder of the paper will seek additional problem constraints that enable performant policies to be learned in resources that grow as a small polynomial in $n$.  We encounter a challenge when this problem structure is examined from the lens of network architectures in deep RL.  The most appropriate structure would be for the network structure to be adapted to the structure of the problem which is not given explicitly.  Although one could perform a search over network architectures for utility~\autocite{zoph2016neural, elsken2019neural,evci2020rigging, sutton1993online}, an architecture-search approaches will require a high complexity either in samples or computation for a fully general solution.  Here, we are interested in characterizing a subset of problems where a more efficient search is possible.

The problem setting can be described more formally with the notation from the above section.  Given a set of combinatorially large environments that are parameterized by $n\in \mathbb{N}$, we want algorithms $\mathcal{L}(\theta)$ with resources and hyperparameters given by $\theta(n)$, so that the algorithm $\mathcal{L}(\theta(n))$ is applied to experience from an environment of that scale, $\mathcal{E} \sim \mathcal{E}(n)$.  

We consider the performance of algorithms on the environments as we increase $n$.  The environments can be calibrated 
so the time to reach a threshold level of performance gives good guidance on how the algorithm performs across a wide range of environment scales. 
Suppose we are given a set of environments $\env(n)$, and interface-compatible instances of the learning algorithm, $\lrn(n)$.  We use the average reward to define a performance metric, namely the minimum time to reach a threshold level of performance. 
\begin{equation}
\mathrm{TimestepsToThreshold}(n) = \arg\min_t  r_{\mathrm{thresh}} \leq \mathbb{E}_{\env(n),\lrn(n)} \{R^\env_{t'}\},\ \forall t' \geq t 
\end{equation}

If we double the observation dimension (which squares the number of states in the combined MDP), what happens to the time to threshold performance for different algorithms? 
Define a metric that characterizes how the algorithm performs when the problem dimension is doubled. 
\begin{equation}
\mathrm{TimestepDoublingRatio}(n) = \frac{\mathrm{TimestepsToThreshold}(2n)}{\mathrm{TimestepsToThreshold}(n)} 
\end{equation}
We are interested finding situations where the doubling ratio is nearly constant over some range of $n$, which suggests that the learning complexity is not increasing directly with the state space size.  For algorithms having constant resources on problems of increasing complexity, we would expect this quantity to increase to infinity for some fixed $n$.  In practice, several deep RL algorithms that are developed with a fixed number of hidden units can also fail to learn to an acceptable level of performance when given large unstructured observation vectors.  For algorithms that are able to reliably learn performant policies on large problems, they may vary greatly on the number of computational resources that they require as they scale.  Echoing the sentiment of the bitter lesson~\autocite{bitterLesson}, we can use the computational requirements as a secondary evaluation criteria among algorithms that exhibit reliable scaling.

\section{Towards a Scalable Algorithm}

The combined MDP construction provides a mechanism to define scalable problems, but it does not suggest a mechanism for efficient solutions when the generating process is unknown.  For a general problem that does not expose any statistically-exploitable structures, we should not expect a sample efficient solution. Therefore, we focus on the problem regime of interest, namely problems where there is some computationally accessible statistical structure in the stream of experience that enables sample efficient learning.  Our particular choice of statistical structure is motivated in part by the ideas of predictive state representations~\autocite{littman2001predictive} which showed that predictions can provide an adequate state representation  in theory.  Subsequent work showed the possibility of making many predictions in parallel~\autocite{modayil2014multi}, finding low-dimensional linear embeddings of many predictions~\autocite{boots2011closing}, and using an additional set of predictive questions to improve neural network feature learning through the auxiliary task effect~\autocite{jaderberg2016reinforcement}.  Moreover, several RL works have shown the use of predictions for the construction of reward models and transition models for planning~\autocite{parr2008analysis,sutton2008dyna},  planning-like knowledge transfer~\autocite{dayan1993improving,barreto2017successor} and representation learning~\autocite{Zheng2021LearningSR,martin2021,Schlegel_2021}.  These works raise the possibility that learning algorithms can extract relevant predictive structure directly from the statistics of experience, instead of requiring the problem structure to be specified as part of the problem input. 

We continue this approach, by developing an algorithm that forms its own predictive questions as general value functions, learns approximate answers to these predictive questions, and refines the representations formed by the answers to improve performance.  Notably, all of the processes in the presented algorithm can operate successfully in a fully incremental algorithm, and with sub-exponential computational and sample complexity in practice.

\subsection{General value functions}
Before turning to the algorithm itself, our solution uses general value functions~\autocite{sutton2011horde} to define and answer subproblems, so we first review them.  A general value function prediction is defined by a tuple, $(C,\gamma,\pi)$; with a time varying cumulant signal, $C_t \mathequiv C(S_t)$, a (potentially non-constant) discount factor $\gamma\in[0,1)$, and a policy $\pi$ (limited in this paper to the behavior policy $\pi_b$).  This generalizes the usual value function in reinforcement learning in several ways, most visibly by choosing signals other than the reward to define the objective.  To make this correspondence clear, define the return at time $t$ as the discounted sum of future cumulants while following the policy,
\begin{equation}
    G^{C,\gamma,\pi}_t \mathequiv C_{t+1} + \gamma C_{t+2} + \gamma^2 C_{t+3} + \ldots,
\end{equation}
and then define the general value function for states,
\begin{equation}
    v^{C,\gamma,\pi}(s)  \mathequiv \mathbb{E}\{G^{C,\gamma,\pi}_t \ |\  S_t=s, A_t\sim \pi(S_t), S_{k+1}\sim T(S_k, A_k), A_{k+1} \sim \pi(S_{k+1}) \quad \forall k\geq t \},
\end{equation}
and finally the general value function for actions,
\begin{equation}
    q^{C,\gamma,\pi}(s,a)  \mathequiv \mathbb{E}\{G^{C,\gamma,\pi}_t\ |\ S_t=s, A_t=a, S_{k+1}\sim T(S_k, A_k), A_{k+1} \sim \pi(S_{k+1}) \quad \forall k \geq t\}.
\end{equation}

Many temporal-difference learning algorithms are available for estimating these quantities online from experience due to their formal similarity with conventional value functions.  In particular, when using on-policy learning, the TD(0) algorithm can be applied directly, with different guarantees available depending on the class of function approximation used.  For linear function approximation, we can write the estimate of the GVF from states as $\hat V(\xv;w) \mathequiv w^\top \xv \approx v^{C,\pi,\gamma}(s)$,  where $\xv$ is the feature vector associated with the state $s$.  For function approximation with a differentiable function $f$ and parameters $w$, and a fixed mapping $\psi:\mathcal{O}\rightarrow \Re^m$ from a raw observation $o$ into a base feature vector, we can write the estimate as $\hat V(o;w) \mathequiv f(\psi(o);w)$.  In both cases, we can use standard reinforcement learning methods~\autocite{sutton2018reinforcement}.

\begin{figure}
\centering
(a)
\includegraphics[height=1in]{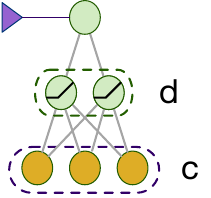}
\hfill
(b)\includegraphics[height=1in]{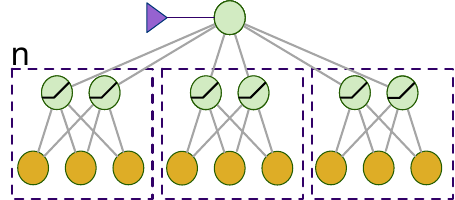}
\hfill
(c)\includegraphics[height=1in]{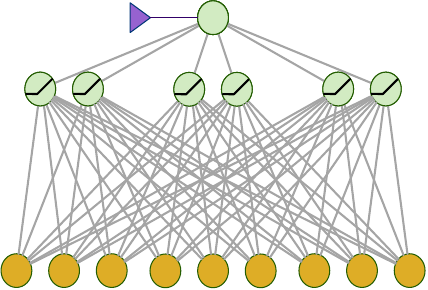}
\caption{Standard solution approaches. The figures show (a) a simple MLP for a single component MDP with $c$ inputs (in orange) and a hidden layer of size $d$ with a nonlinearity connected to a value function estimate (in light green) which is refined using backpropagation by the external value function objective (in the purple triangle); (b) a solution for a combined MDP with $n$ components where the structure of the inputs is externally given; and (c) a default fully connected architecture for when the architecture designer does not have the problem structure a priori.  The fully connected architecture covers the function space of the combined MDP, but it requires substantially more resources and may exhibit worse sample efficiency.}
\label{fig:one-component}
\end{figure}

\subsection{Standard architectures}

Before considering new algorithms for this problem setting, we review the form of conventional function approximation architectures for such problems, in the situation where the structure is known or unknown.
First consider a single component MDP, when we know an standard multi-layer perceptron (MLP) is sufficient for finding accurate action values (and an epsilon-greedy policy) for a single component.   On the left of Figure~\ref{fig:one-component}, we show an example of a flat network with $c$ inputs and $d$ hidden units that can reliably learn a performant policy on an MDP with a single component.  The external objective is shown with a triangle, that provides feedback to a node representing the answer from the network.  In the center diagram, we show how a set of small networks can be combined for learning on an MDP with multiple components when the structure of the inputs is known, i.e. the person defining the network knows how to partition the input into useful components.  When the structure of the input is unknown prior to network definition, then we have the situation on the right of Figure~\ref{fig:one-component}.  Here, every hidden unit is combined with every base feature from every component to cover the original function space in the middle.  This approach of connecting to every unit has a clear cost in memory resources (for storing the weights) and in compute (for computing all the weight-activation products), and potential costs for efficient credit assignment and training time.

\subsection{Nibbler architecture}
The idea we pursue is to enable the algorithm to define its own components by using GVF predictions to construct subproblems.  

\begin{figure} {
    \begin{minipage}{3in}
    \centering
    (a)
    \includegraphics[height=1in]{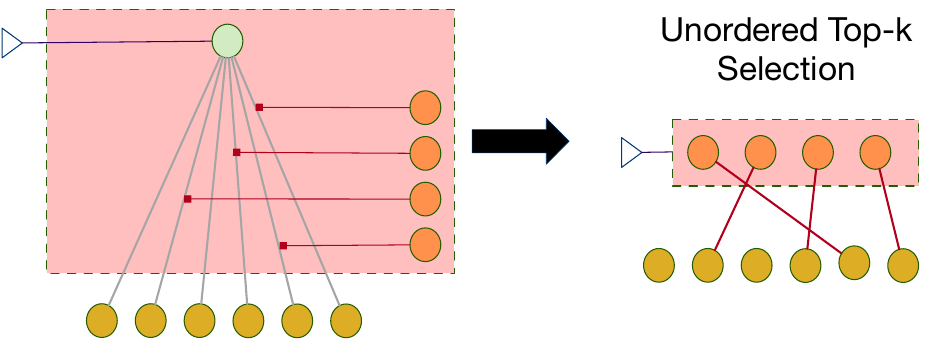}
    (b)
    \includegraphics[height=1in]{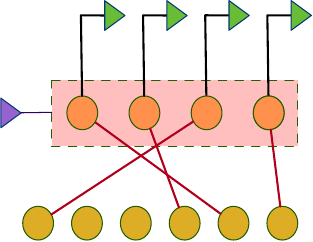}
    \end{minipage}
\begin{minipage}{3in}
(c) \includegraphics[height=2in]{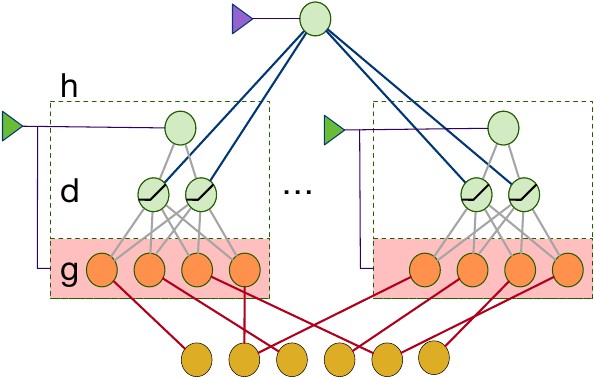}
\end{minipage}
   }
    \caption{On the top left (a) we show the process of selecting the top $k$ features, based on a measured utility for each base feature.  The utility measure examined in this work is the absolute value of the weights in a linear GVF answer.  The implementation in the appendix performs a lazy linear-computation selection process, where at most one selection is swapped on every call.  On the bottom left (b) we show that new GVF questions (green triangles) can be defined for each selected feature.  In this work, we define each GVF question by setting the GVF cumulant to be the feature, with the same discount as the main problem, and an on-policy prediction question.  The input GVF question is the one step reward (the cumulant is set to the external reward with $\gamma=0$).  On the right (c) is the core of the Nibbler architecture (shown without the linear skip connection to the base features).}
    \label{fig:nibbler}
\end{figure}

We start with an operation that selects a set of $k$ features based on the utility of individual features in a GVF answer for a GVF question.  
A simple definition of utility is the magnitude of the weight associated to a feature in a GVF answer that uses linear function approximation.  Suppose we have a GVF question whose GVF answer, $\hat V$, is approximated by the dot product of the feature vector $\xv$ and weight vector $\wv$, $\hat V=\sum_{i=1}^{m} \wv[i] \xv[i]$ (we use $m\mathequiv nc$ as the total number of base features).  Define the utility of feature $i$ for this question as $|\wv[i]|$.  Select the top $k$ features by utility, storing them as a list of indices $L\subset \{1,...,m\}$. This produces a new feature vector $\xv[L]$ of length $k$.  We note that process of top-$k$ selection can be performed in an incremental manner (shown in the appendix), where at most one index is changed per call.  Moreover, this is possible using only linear memory and compute, yielding a computational cost of $O(n)$.  This process of selecting the top-k features is shown in Figure~\ref{fig:nibbler} (top left).  We note in passing that many other utility measures could be defined over the features in a GVF answer.

The above feature utility is used in two ways in the Nibbler algorithm; selecting useful features for linear reward prediction as cumulants in constructed GVF questions, and selecting useful features for a GVF answer as base features for constructing new   nonlinear features.

We first use the feature selection process for GVF question construction, as shown in Figure~\ref{fig:nibbler} (bottom left).  The input GVF question is a one timestep prediction of external reward, $\text{GVF}^\text{Reward} \mathequiv (R_{t+1}, 0, \pi_b)$, for which a linear answer is learned over the base feature vector from the observations.   The top $h$ features are used to define new GVF questions (shown with green triangles in the figure).  These GVF prediction questions use the top features as cumulants, $C^i_{t+1} \mathequiv \xv_{t+1}[L[i]]$, and use the external discount and behavior policy, $\text{GVF}^i \mathequiv (C^i_{t+1}, \gamma, \pi_b)$.  This procedure provides an algorithm for defining new GVF questions directly from the external reward signal, where the number of internally constructed questions, $h$, is a hyperparameter that can be defined based on the problem size, $n$.  This raises the possibility of using an internally specified GVF question to define new GVF questions, by a mechanism that is directly influenced by external reward, but that goes beyond the scope of the current work.

These $h$ GVF questions are used to define components and construct new nonlinear features, as shown in Figure~\ref{fig:nibbler} (right).  The same feature selection process is used to identify the top $g$ base features for each GVF subproblem.  In component $i$, the selected base features are used as inputs for a MLP generating a vector of $d$ nonlinear features, $\xv^{\text{e},i}$, as part of a (linear) answer to the question $\text{GVF}^i$.  This MLP generates action value estimates that are not used directly---the entire role of the MLP is to construct useful nonlinear features for answering the subproblem.  Finally, all the nonlinear features from all the components, along with the original base features, are concatenated to form a single large feature vector, $\xv^\text{f}$. This full feature vector is used for the externally defined problem, $\text{GVF}^\text{ext} \mathequiv (R_{t+1}, \gamma, \pi_b)$, where action values are learned on-policy with linear function approximation.  

We call the combined algorithm Nibbler, due to the piece-wise manner in which the algorithm operates.  Some additional algorithmic details are useful to highlight.  All parts of the algorithm can run fully incrementally online, without distinct phases.  Although all the steps so far have not modified the policy, we obtain policy improvement through epsilon-greedy action selection, much like in the conventional use of SARSA for control.  The algorithmic form we use for estimating action values is QV~\autocite{wiering2009qv}, which has the same objective as expected SARSA but bootstraps directly from state value estimates.  Thus, we use only state value estimates for both feature selection processes, but we use action value estimates for the construction of nonlinear features and for control.  When estimating action values with a neural network, we follow the common practice of having the action values for a single GVF question share the hidden layer with the state values.  We use SGD with momentum as the step-size optimizer to adapt the weights as described in the appendix.

 \subsection{Computational Complexity}
 
 We can coarsely characterize the memory and compute costs by the number of weights in these networks, starting with the flat baselines in Figure~\ref{fig:one-component} (center and right).  Say there are $n$ components, each with $c$ base features and learnable with $d$ hidden features.
 The number of weights is dominated by connection between the inputs and the hidden layer, so we focus on it.  There are $O(ndc)$ weights for the network with known structure, and $O(n^2dc)$ weights for the flat network for unknown structure.

 The diagram in Figure~\ref{fig:nibbler} shows the main components of the Nibbler algorithm.  The selection of base features for GVF cumulants (lower left of figure) operates on the base features, so has a complexity of $O(n)$ per timestep.  The explicit selection of $g$ features within each of the the $h$ components can greatly reduce the complexity of the connections from each hidden unit from $nc$ to $g$, yielding a complexity of $O(ghd)$ for the middle portion. This is favorable in comparison to the $O(n^2dh)$ complexity of the flat architecture above, when $n\approx h$ and $c \approx g$.  The remaining complexity comes from the selection of $g$ base features for each of the components, which has a cost of $O(hnc)$.  When $n$ grows large, the quadratic cost of selecting features can start to dominate the costs of the main learning process.  A practical mitigating factor for the quadratic complexity of this piece of the algorithm is the efficient parallel implementation on modern hardware accelerators.

\begin{algorithm}
\caption{The Nibbler algorithm}
\begin{algorithmic}[1]
\Require $\psi: \mathcal{O} \rightarrow \{0, 1\}^m$ be the $m$ base features from the observation
\Require $z$ the number of discrete actions, $\mathcal{A} \mathequiv \{ a_1,\ldots, a_z \}$
\Require $\epsilon$ the probability of taking a random action 
\Require $\gamma$ the discount for the GVF nexting questions and the main problem
\Require $h$ the number of GVF nexting questions in QV form
\Require $d$ the number of non-linear features produced by each GVF answer
\Require $g$ the number of base features used in each GVF answer
\Statex
\Statex Let  $\hat{Q}$ be the action value estimates for the main learner, initialized to zero, learned in QV form
\Statex Let $I \mathequiv \{1,\ldots,h\}$ be the indices for the GVFs
\Statex Let  $C_L :\{1,\ldots, h\} \rightarrow \{1,\ldots,m \}$ select $h$ base feature indices for GVF cumulants (as a list of length $h$)
\Statex Let  $K^i_L: \{1,\ldots,g \} \rightarrow \{1,\ldots, m\}$ select $g$ base feature indices for GVF answer $i \in I$
\Statex Let   $K^i: \Re^m \rightarrow \Re^g$ be the associated linear projection ($K^i(\xv) \mathequiv \xv[K^i_L]$)
\Statex Let  $\phi^i: \Re^g \rightarrow \Re^d$ be the non-linear features constructed for GVF answer $i \in I$
\Statex
\While{true}
\State Receive reward $R_{t+1}$ and observation $O_{t+1}$ from the environment
\State $\xv^{\text{b}}_{t+1} \leftarrow \psi(O_{t+1})$
\State $\forall i \in I,\  \ \xv^{\text{e},i}_{t+1} \leftarrow  \phi^i( K^i(\xv^{\text{b}}_{t+1}))$
\State $\xv^{\text{f}}_{t+1} \leftarrow \text{concatenate}(\xv^\text{b}_{t+1}, \text{concatenate}(\xv^{\text{e},i}_{t+1}, \forall i \in I))$
\State $A_{t+1} \leftarrow $ Select $\epsilon$-greedy action from $\hat{Q}$ (linear on feature vector $\xv^{\text{f}}_{t+1}$)
\State Send action $A_{t+1}$ to the environment
\State Update parameters of $\hat{Q}$ using reward $R_{t+1}$
\State Update parameters of GVF answer $i$, $\forall i \in I$
\State Update base feature selection for GVF answers, $K^i_L\,  \forall i \in I$
\State Update base feature selection for cumulants of GVF questions, $C_L$ 
\EndWhile
\end{algorithmic}
\end{algorithm}

\section{Algorithm Pseudocode}

 \begin{algorithm}
 \caption{Update main state value and action value estimates (Algorithm 1: line 8)}
 \begin{algorithmic}[1]
\Require $\text{co-opt}_\Lambda$: a step-size gradient co-optimizer with hyperparams $\alpha_\Lambda$ and params $\vartheta_\Lambda$ for names in $\Lambda$
\Statex Let $\hat V(\xv^\text{f}; w_\text{v}) \mathequiv (w_\text{v})^\top \xv^\text{f}$ 
\Statex Let $\hat Q(\xv^\text{f},a; w_{\text{q},a}) \mathequiv (w_{\text{q},a})^\top \xv^\text{f}$ 
\State $ Y \leftarrow \text{stop-gradient}(R_{t+1} + \gamma \hat V(\xv^{\text{f}}_{{t+1}}; w_\text{v}))$
\State $L \mathequiv \frac{1}{2} \left(  ( Y - \hat V(\xv^{\text{f}}_{{t}}; w_\text{v}) )^2  +  ( Y - \hat Q(\xv^{\text{f}}_{{t}}, a_t); w_\text{q,a}) ^2 \right) $
\For{ $\Lambda \in \text{v}, (\text{q},a_t)$}
\State $w_\Lambda, \vartheta_\Lambda \leftarrow \text{co-opt}_\Lambda( w_\Lambda, \alpha_\Lambda, \frac{\partial L}{\partial w_\Lambda} , \vartheta_\Lambda)$  
\EndFor  
\end{algorithmic}
 \end{algorithm}

\begin{algorithm}
 \caption{Update weights for GVF answers (Algorithm 1: line 9)}
 \begin{algorithmic}[1]
   \Statex Let $w^{i}_{\text{v}},w^{i}_{\text{q},a}  \in \mathbb{R}^d$ be weights for state and action value estimates for GVF question $i \in I$
      \Statex Let $w^{i}_{\text{u}} \in \mathbb{R}^u$ be weights for nonlinear features for the estimate of GVF $i \in I$
\Statex Let $\hat V^i(\xv^{\text{e},i}; w^i_\text{v}) \mathequiv (w^i_\text{v})^\top \xv^{\text{e},i}$ 
\Statex Let $\hat Q^i(\xv^{\text{e},i},a; w^i_{\text{q},a}) \mathequiv (w^i_{\text{q},a})^\top \xv^{\text{e},i}$ 
\Statex Let $\xv^{\text{e},i} \mathequiv 
\phi^i(\xv^{\text{b},i}) = 
\phi(\xv^{\text{b},i}; w^{i}_\text{u}) $ the features from the $i$-th GVF, where $\phi$ is a differentiable function.
\Statex Let $\xv^{\text{b},i}\mathequiv 
K^i(\xv^{\text{b}}) = \xv^\text{b}[K^i_L] $ the non-differentiable selection of inputs to the $i$-th GVF answer.
\For{$i \in I$}
 \State $C^i_{t+1} \leftarrow \xv^\text{b}_{t+1}[C_L[i]] $
 \State $ Y \leftarrow \text{stop-gradient}(C^i_{t+1} + \gamma \hat V^i(\xv^{\text{e},i}_{t+1}; w^i_\text{v}))$
\State $L^i \mathequiv \frac{1}{2} \left(  ( Y - \hat V(\xv^{\text{e},i}_{t}; w^i_\text{v}) )^2  +  ( Y - \hat Q(\xv^{\text{e},i}_{t}, a_t; w^i_{\text{q},a_t}) ^2 \right) $
\For{ $\Lambda \in \text{v}, (\text{q},a_t),  \text{u}$}
\State $w^i_\Lambda, \vartheta^i_\Lambda \leftarrow \text{co-opt}^i_\Lambda(  w^i_\Lambda, \alpha^i_\Lambda, \frac{\partial L^i}{\partial w^i_\Lambda} , \vartheta^i_\Lambda)$  
\EndFor  
\EndFor
\end{algorithmic}
 \end{algorithm}
 
 \begin{algorithm}
\caption{Update $K^i_L$: feature selection for answers (Algorithm 1: line 10)}
\begin{algorithmic}[1]
  \Require $\alpha_{\text{b}}$ a stepsize
\Require $K^i_M \in \{0,1\}^m$ to indicate entries in $K^i_L$ (i.e.  $K^i_M[j]=1 \iff j\in K^i_L$)
  \Statex Let $w_\text{support}^{i} \in \mathbb{R}^m$ be weights for a state value function answer for GVF question $i \in I$
   \For{$i\in I$}
   \State $  K^i_L, K^i_M, changed, pos \leftarrow \text{incremental-top} ( K^i_L, K^i_M, \tau_K,  | w_\text{support}^{i} | ) $
\State if $changed$ then reinitialize GVF weights for input $pos$ (both $w_\text{u}^{i,pos, \cdot}$ and $\vartheta_\text{u}^{i,pos, \cdot}$ )
\State $ \delta \leftarrow C^i_{t+1} + \gamma  (w_\text{support}^i)^\top \xv^\text{b}_{t+1} - (w_\text{support}^i)^\top \xv^\text{b}_{t} $
\State $w_\text{support}^{ i}  \leftarrow w_\text{support}^{ i} + \alpha_{\text{b}} \delta \xv^\text{b}_{t}$
\EndFor
\end{algorithmic}
\end{algorithm}

\begin{algorithm}
\caption{Update $C_L$: Feature selection for question cumulants (Algorithm 1: line 11)}
\begin{algorithmic}[1]
  \Require $\alpha_{\text{b}}$ a stepsize
\Require $C_M \in \{0,1\}^m$ to indicate entries in $C_L$.
  \Statex Let $w_\text{discovery} \in \mathbb{R}^m$ be weights for a state value function answer for the main reward.
\State $C_L, C_M,  changed, pos \leftarrow \text{incremental-top} ( C_L,  C_M,\tau_I, |w_\text{discovery}|) $
\State if $changed$ then reinitialize GVF weights to GVF question $pos$ ($w^{pos}_\text{u}$ and $\vartheta^{pos}_\text{u}$)
\State $ \delta \leftarrow R_{t+1} - (w_\text{discovery})^\top \xv^\text{b}_{t} $
 \State$ w_\text{discovery}  \leftarrow w_\text{discovery} + \alpha_{\text{b}} \delta \xv^\text{b}_{t}$
 \end{algorithmic}
\end{algorithm} 

  The main algorithm is shown as Algorithm 1.  Lines 2-3 perform a given transformation of the observation into base binary features.  Line 4 computes the nonlinear features generated by each GVF answer, and line 5 concatenates all the non-linear features and the base features into a single feature vector for the main control learner.  Lines 6-8 correspond to epsilon-greedy action selection from the linear action value estimate over the full feature vector, followed by an on-policy update of those parameters.  Line 9 performs deep RL, to update each on-policy GVF answer.  Line 10 updates the selection of $g$ base features for each GVF answer, and line 11 updates the selection of $h$ features to be cumulants for the GVF questions.

For the construction of GVF questions, we form a one-step linear reward model in a problem's given base features.  Every feature that is relevant for predicting reward is identified and is used to form a separate GVF question, by using the feature activation as a cumulant, the original problem's discount, and a policy which is the behavior policy (on-policy prediction learning).

\begin{table}[h]
    \centering
    \begin{tabular}{c|c | l}
    symbol & value & description\\  \hline
         $h$ & $2n$ & number of gvf questions (twice the number of boards)\\
         $g$ & 82 & base features per GVF answer  \\
         $d$ & 256 & nonlinear features per GVF answer  \\
         $\alpha$ & $\kappa /\sqrt{h}$ & stepsize for main board \\
         $\tau$ & 0 & threshold for swaps in top-k selection \\
         $\gamma$ & 0.99 & discount factor \\
         $\alpha_b$ & $\kappa/\sqrt{h}$ & step size for linear learners\\
         $\kappa$ & $0.001\sqrt{2}$ &  domain-specific step size factor \\
         $\nu$ & 0.99 & momentum for all learners
     \end{tabular}
    \caption{Key hyperparameters}
    \label{tab:my_label}
\end{table}

\section{Domain: Multi-catch Environments}

\begin{figure}
\centering
\includegraphics[width=.8\textwidth]{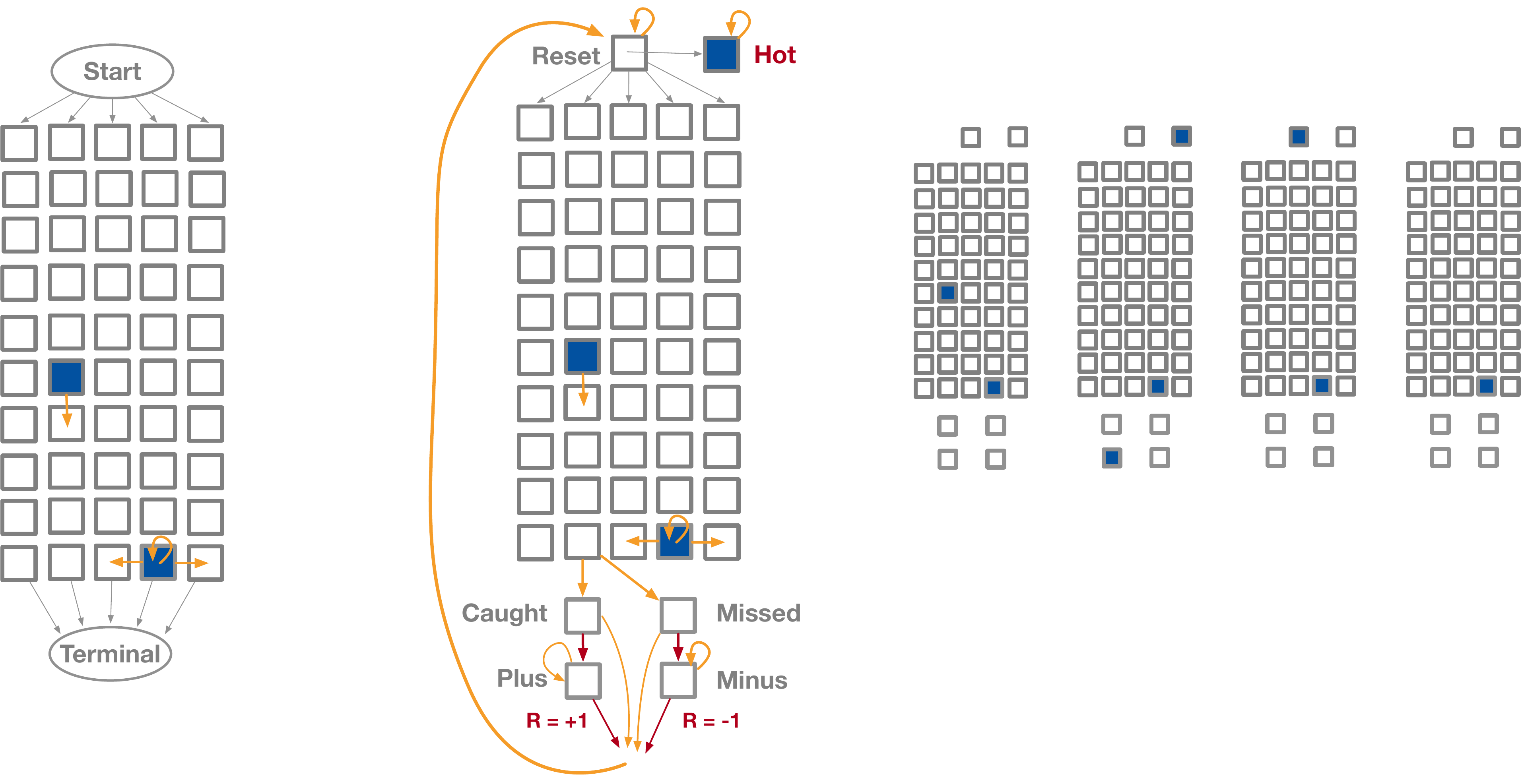}
\caption{A standard episodic Catch domain (left), the transformation of the standard catch domain into a continuing problem (center), and an observation from a multi-catch domain that is composed of four catch boards (right). The domain structure may not be directly accessible through the observation (the observation bits can be permuted), and the constituent boards may have no shared structure or dynamics.}
\label{fig:multi-catch-domains}
\end{figure}

We now develop multi-catch, a family of combinatorial environments that allow flexible manipulation of the problem scale. 
We construct a multi-catch environment by combining multiple instances of a variation of the catch problem, with the number of instances determining the problem scale. 
To fully describe a multi-catch environment, we first review the common catch problem, next we introduce a continuing version of the originally episodic catch problem, finally we discuss how instances of continuing catch are combined into a composite multi-catch environment.

Episodic catch, depicted in Figure \ref{fig:multi-catch-domains} (left), involves moving a paddle on a two-dimensional board with the goal of intercepting a falling ball~\autocite{Osband2019}. 
The observation consists of a set of 50 bits, commonly arranged in a rectangular board of 10 rows and 5 columns.
On the bottom row is the paddle with a location indicated by an active bit.
The paddle can be moved horizontally by one position using three categorical actions (called left, right, and stay).
At the first timestep of the episode, the ball arrives in the top row in one of the columns, also indicated by an active bit.
The ball falls by one row at a time, until it arrives at the bottom row, after which a non-zero reward is emitted and the episode terminates. 
If the position of the ball and paddle are the same for this last observation, the reward is +1, otherwise it is -1.

First, we create a continuing version from the episodic catch by adding a reset bit.
The ball now returns to the reset bit when the episode would normally end.
When in the reset bit, the ball may transition back into the board, with a small probability.
We make the events in which the ball is caught or missed observable by adding a catch and miss bit.
The catch bit is active in the timestep after the ball is caught and inactive otherwise.
Likewise, the miss bit is active in the timestep after the ball is missed and inactive otherwise.

Next, we augment the system with the hot bit (Section \ref{sec:problem-formalization}), determining whether the a non-zero reward will be emitted after the ball is caught or missed.
When the hot bit is off, catching or missing the ball emits zero reward.
A board becomes hot with a small probability each time the ball enters the board, and it continues to stay hot until the ball exits and a nonzero reward is emitted.

Finally, we add bits that make the forthcoming reward observable: a plus bit that becomes active the timestep after the catch bit is on if the board is hot, and a minus bit that becomes active the timestep after the miss bit is on, again, if the board is hot.
These bits stay active for a stochastic duration and are followed by a reward once they get inactive; the plus bit is followed by a reward of +1, whereas the minus bit results in a reward of -1.
As the reward is emitted, the ball returns to the reset bit.
The observation includes the hot, reset, catch, miss, plus, and minus features and the features for the ball and paddle positions. 
We see this augmented version in Figure \ref{fig:multi-catch-domains} (center).

We are now ready to describe how we combine the base catch problems into a composite environment.
In the composite environment, the balls arrive independently in the individual boards, and can potentially be present simultaneously; rewards from the individual boards are summed to obtain the joint reward signal.
Independent ball dynamics and the form of the joint reward signal lends the composite multi-catch environment its factorial structure: the transition probabilities and the reward function can be factored into those of the component catch MDPs.

The action space, on the other hand, is coupled and the same action is broadcast to each component catch.
This constraint limits the action space to a set of three discrete choices (left, right, and stay), with the same action moving paddles in the constituent boards jointly in the intended direction.  To prevent the paddles in all components moving identically, we introduce stochasticity into consequences of the selected action, so each paddle ignores the requested action and moves independently at random with the probability $\text{paddle\_noise}$.

The observation spaces is formed by concatenation of the individual observation features from the component catch boards, which then undergoes a fixed random permutation before presentation to the learning algorithm.
An observation from a multi-catch domain with four boards is shown in Figure \ref{fig:multi-catch-domains} (right). 
The environment settings to instantiate a multi-catch environment are described in full in Table \ref{tbl:multi-catch-config}.

In multi-catch, sample-efficient learning is possible despite its combinatorial complexity.
Learning to perform well in a multi-catch domain is essentially learning to catch in the component catch boards.
As the dynamics are factored, for predicting a feature in a given component, only the features in the same component are relevant; learning about the individual boards, in principle, can proceed in parallel and independently (subject to resource constraints). 
This suggests that learning a good policy in multi-catch can be efficient: not requiring a number of samples that is exponential in the number of components.

\begin{table}
\centering
Environment parameters that vary in main experiments
\begin{tabularx}{\linewidth}{ 
  | l | c 
  | >{\arraybackslash}X | } 
  \hline
 \textbf{Environment Setting} & \textbf{Range} & \textbf{Description} \\ \hline
 num\_parallel & 2--128 & Number of parallel catch boards in a multi-catch instance. \\
 \hline
   $p_{hot}$ & 2/num\_parallel &  Probability of becoming hot when the ball enters the board. \\ \hline
   \end{tabularx}

\vspace*{0.5cm}
Environment parameters that were held constant in main experiments
\begin{tabularx}{\linewidth}{ 
  | l | c 
  | >{\arraybackslash}X | }\hline
  \textbf{Environment Setting} & \textbf{Range} & \textbf{Description} \\ \hline
  $p_{arrival}$ & $0.2$ & Probability with which a ball exits the reset bit and enters the board.\\
 \hline 
 $p_{reward}$ & $0.2$ & Probability with which a ball exits the plus or minus bit and returns to the reset bit.\\
 \hline

 paddle\_noise & $0.2$ & Probability that a paddle is moved in a random direction, as opposed to the direction intended by the action. \\ \hline

 num\_rows & $10 $ & Number of rows in the constituent boards \\
 \hline
 num\_cols & $5$  & Number of columns in the constituent boards \\
 \hline
\end{tabularx}
\caption{\label{tbl:multi-catch-config} Multi-catch configuration variables.}
\end{table}

\section{Experiments and Analysis}

We present experimental results on the action-coupled catch environments, for baseline algorithms and the Nibbler algorithm.  The core findings are (1) that standard function methods for RL algorithms fail to perform as well as Nibbler on these environments,  (2) the Nibbler algorithm has a nearly linear relationship between the sample complexity and observation size up to a problem size of 128 on these environments, (3) both Nibbler and a flat architecture improve with the addition of resources---however, beyond a certain problem scale they do not learn effective policies within a comparable amount of experience to what Nibbler requires, and (4) the Nibbler algorithm continues to perform well even when the base environments have somewhat different sizes and dynamics.

\subsection{Nibbler outperforms baseline function approximation architectures}
\begin{figure}
\includegraphics[width=6in]{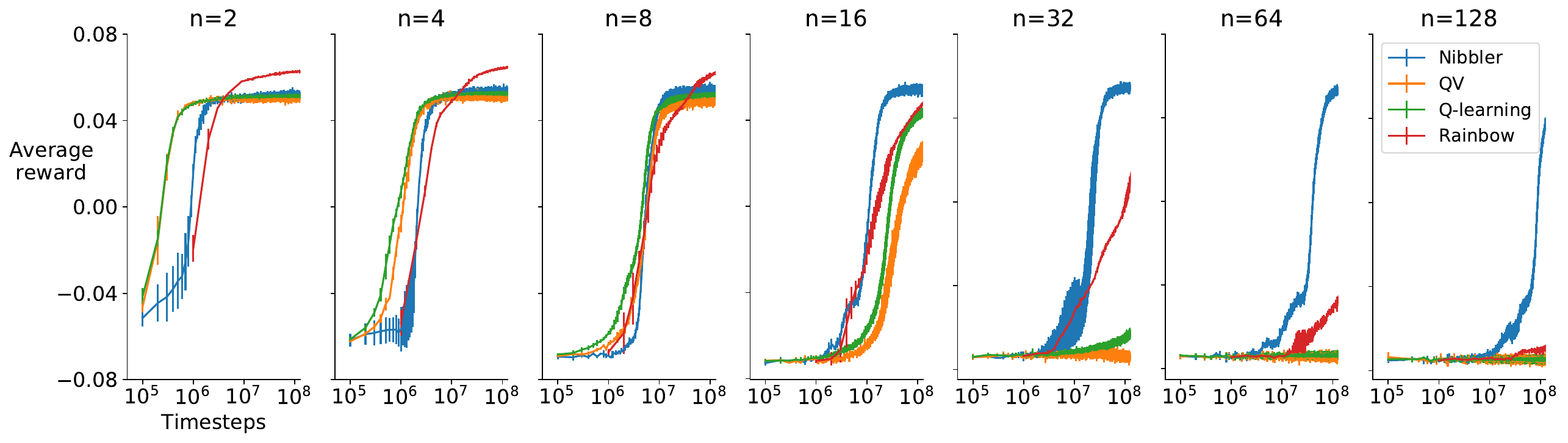}\\
\includegraphics[width=6in]{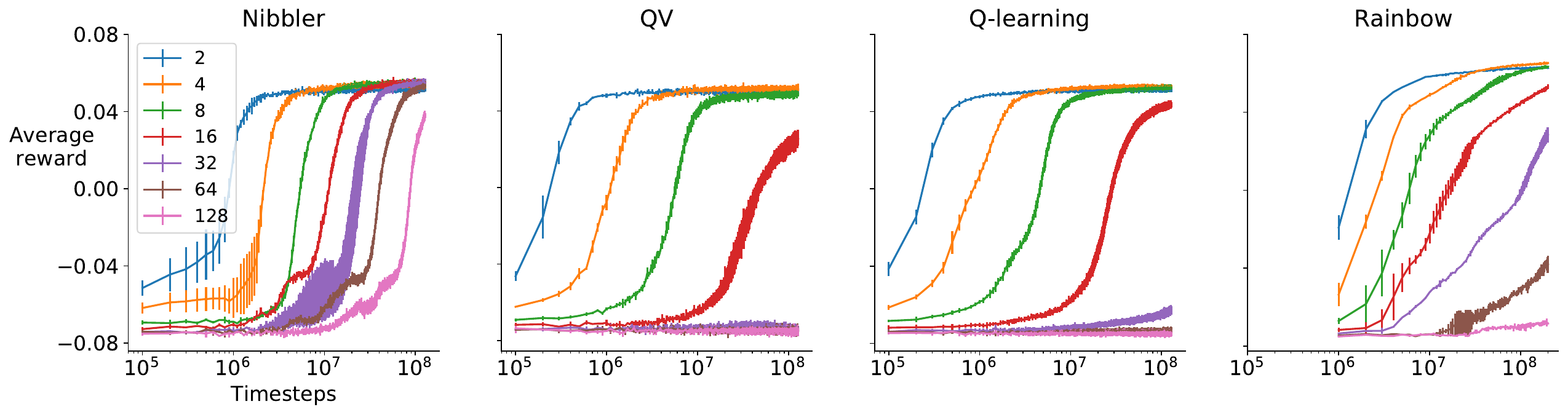}
\caption{\label{fig:scale1}Nibbler outperforms fixed size RL baselines on larger problems.}
\end{figure}

The Nibbler algorithm outperforms baseline RL algorithms with strictly incremental (or conventional deep RL experience-replay) deep function approximation on a parameterized set of problems for large environment sizes, as shown in Figure~\ref{fig:scale1}.  The Nibbler, Q-learning, and QV-learning implementations were fully incremental, with neither target-networks nor experience replay buffers that are commonly used to stablize the learning dynamics of deep RL algorithms. All used a random action selection probability of $\epsilon=0.1$.  Both Q-learning and QV-learning use a hidden layer of size 256, and a SGD learning rate of $\alpha=0.001$ with a momentum of $\nu=0.99$. The Nibbler implementation here uses $d=256$ hidden nonlinear units per question with a stepsize $\alpha=0.001\sqrt{2}/\sqrt{h}$, sets the number of questions to $h=2n$, and selects $g=82$ base features per question.  The Rainbow algorithm combines successful deep RL practices~\autocite{hessel2018rainbow}, and we used an implementation with modifications to support the non-Atari environment~\autocite{dqnzoo2020github}.  Following the original, this implementation does not use epsilon-greedy action selection, but uses internal stochasticity from noisy networks---enabling it to achieve higher asymptotic performance on the smaller problems than the other algorithms.  This implementation selects an action independently for each timestep without the frame-stacking that is common on Atari.  The inputs were passed through a shared layer of 256 hidden units, followed by 512 hidden units per target (state or action value),  followed by the distributional loss, and trained with the default hyperparameters from the paper.

The graphs show that all the implementations learn a good policy up to $n=8$ boards, but then they all exhibit performance drops prior to $n=128$ boards except for Nibbler.  This limitation should not be too surprising, as all the algorithms other than Nibbler have a fixed-sized shared representation, and they offer no systematic way to set learning rates for different sizes.

\subsection{Nibbler's performance exhibits reliable dependence on experience for a range of problem scales}
\begin{figure}
\includegraphics[width=3in]{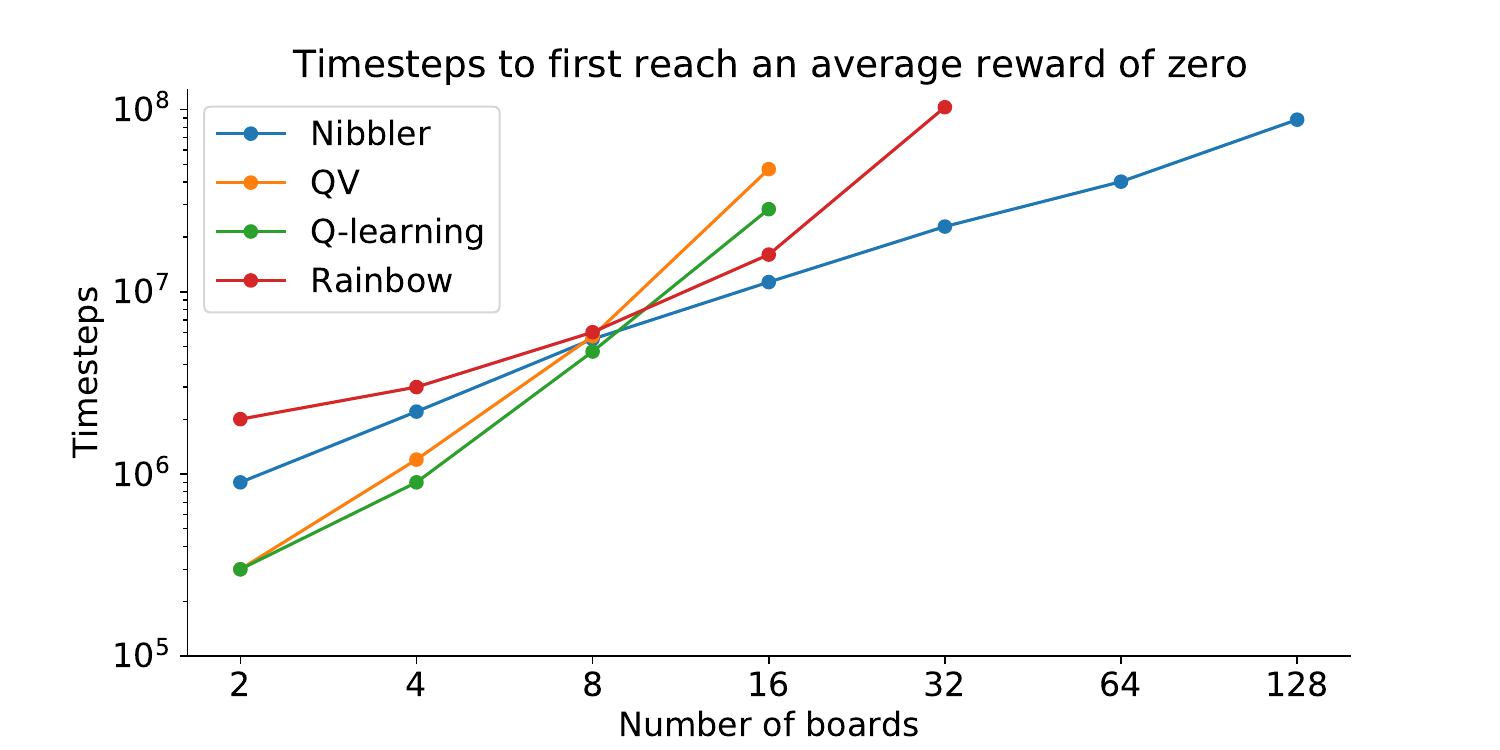}
\includegraphics[width=3in]{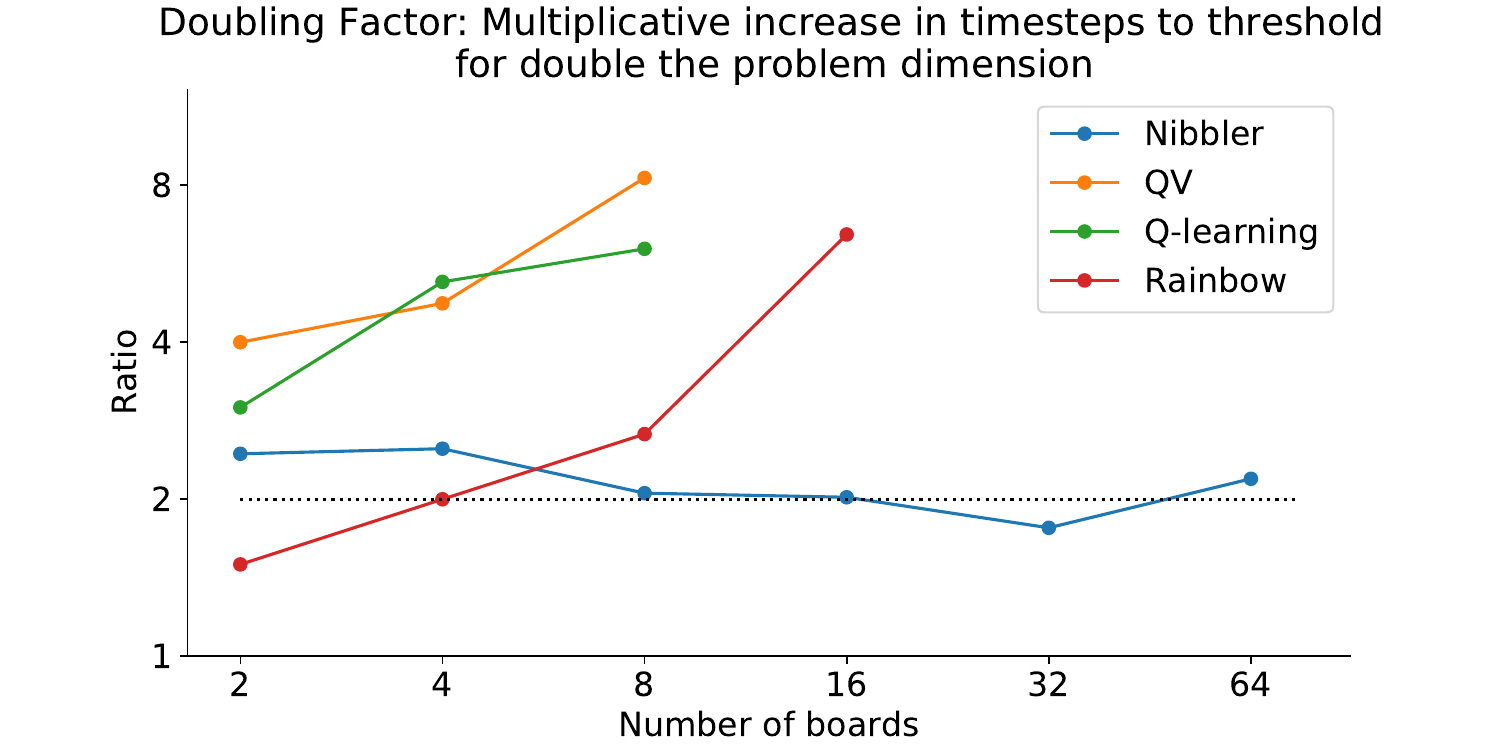}
\caption{\label{fig:nibble-scaling} Nibbler's performance is reliable and scales well across the tested problem sizes.  The number of seeds used varied by algorithm: Nibbler (3), QV (3), Q-learning (10), and Rainbow (2). }
\end{figure}

Building on the earlier result, we examine scaling using the metrics defined in Section 2.3.  On the left of Figure~\ref{fig:nibble-scaling} we plot the TimestepsToThreshold for all the algorithms, and on the right we plot the TimestepDoublingRatio.  We see all algorithms other than Nibbler curving up on the left before reaching 128 boards on the left.  On the right we see that Nibbler requires roughly twice the experience to reach the performance threshold when the number of boards is doubled.  This scaling is only exhibited empirically for the tested ranges.  Other factors may impair performance at larger problem scales than the regime tested here.

\begin{figure}
    \centering
    \begin{tabular}{cc}
    $n=8$ & $n=32$\\
\hspace{-2mm}\includegraphics[height=1.45in]{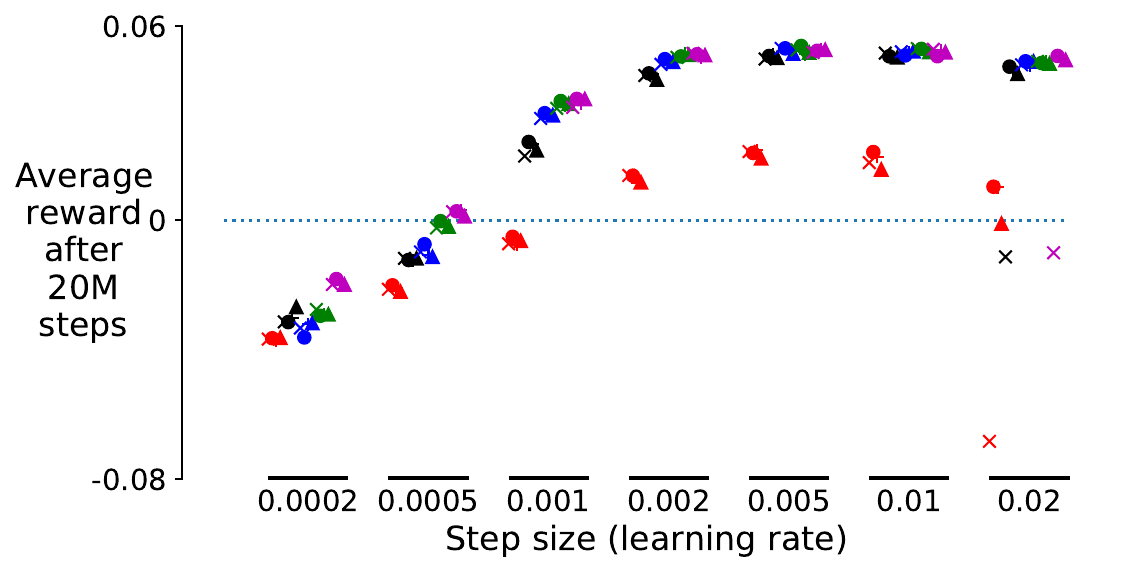}  &
\includegraphics[height=1.45in]{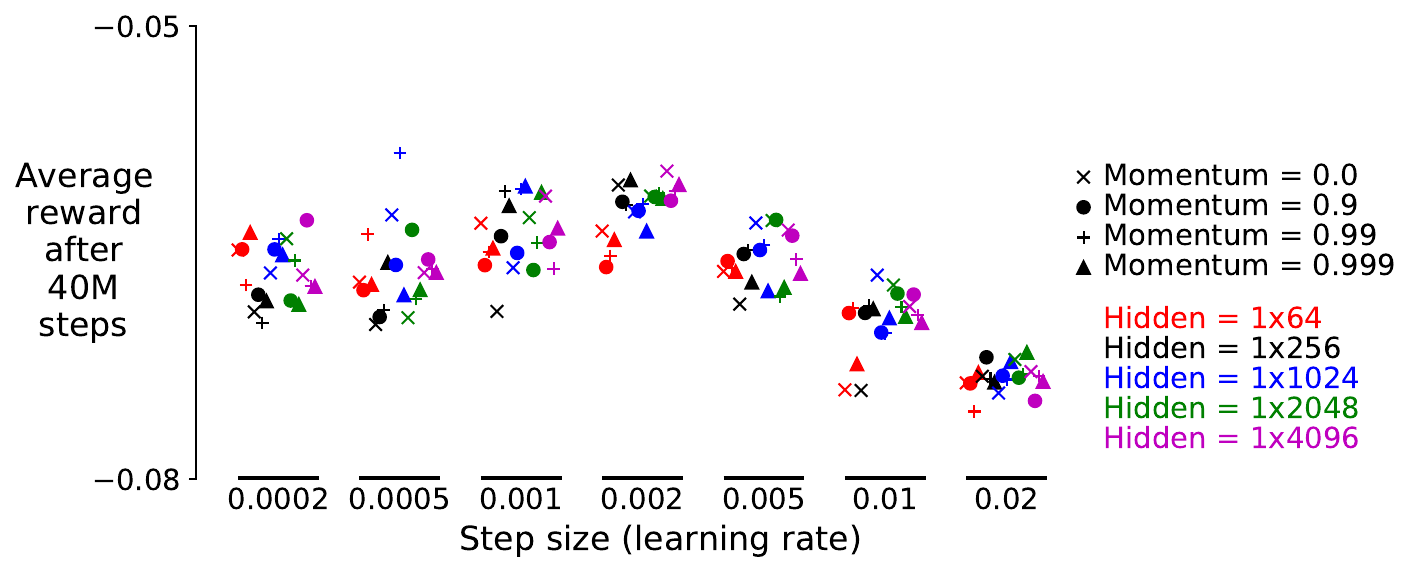} 
\end{tabular}
    \caption{A fully-connected Q-learning agent learns a good policy for an 8 board problem (left) across a range of hyperparameters, but on a 32 board problem (right) it fails to achieve positive average reward in 40 million timesteps for any tested hyperparameter combination. The algorithm was tested across a range of stepsizes, momentum values, and network shapes. No configuration achieved an average reward over -0.05.  In contrast, the Nibbler algorithm reliably achieves an reward over 0.04 at 40M steps (see Figure \ref{fig:scale1}).  The consistent failure supports the argument that increasing computational resources or better hyperparameters with a baseline algorithm is insufficient to learn as quickly as Nibbler on these problems. 
    }
    \label{fig:cc32-q-fail}
\end{figure}

The challenge of understanding how a baseline algorithm performs for a larger $n$ with different hyperparameters motivates the plot in Figure~\ref{fig:cc32-q-fail}.  The experiment swept over a variety of hyperparameter combinations with 2 seeds each, and examined the final performance well after when Nibbler had learned a successful policy (40M timesteps for $n=32$).  On the left we see that a wide variety of hyperparameter choices yield success for Q-learning with a small number of boards ($n=8$).  However, when we increase the number of boards by a factor of $4$ to $n=32$, there is no combination that learns successfully with the same sample complexity of Nibbler, even with a hidden layer that is much larger (up to $4096$ units), across a wide range of step-size settings.  

\subsection{Nibbler and Q-learning performance improves with the number of hidden units}

\begin{figure}
\includegraphics[height=1.5in]{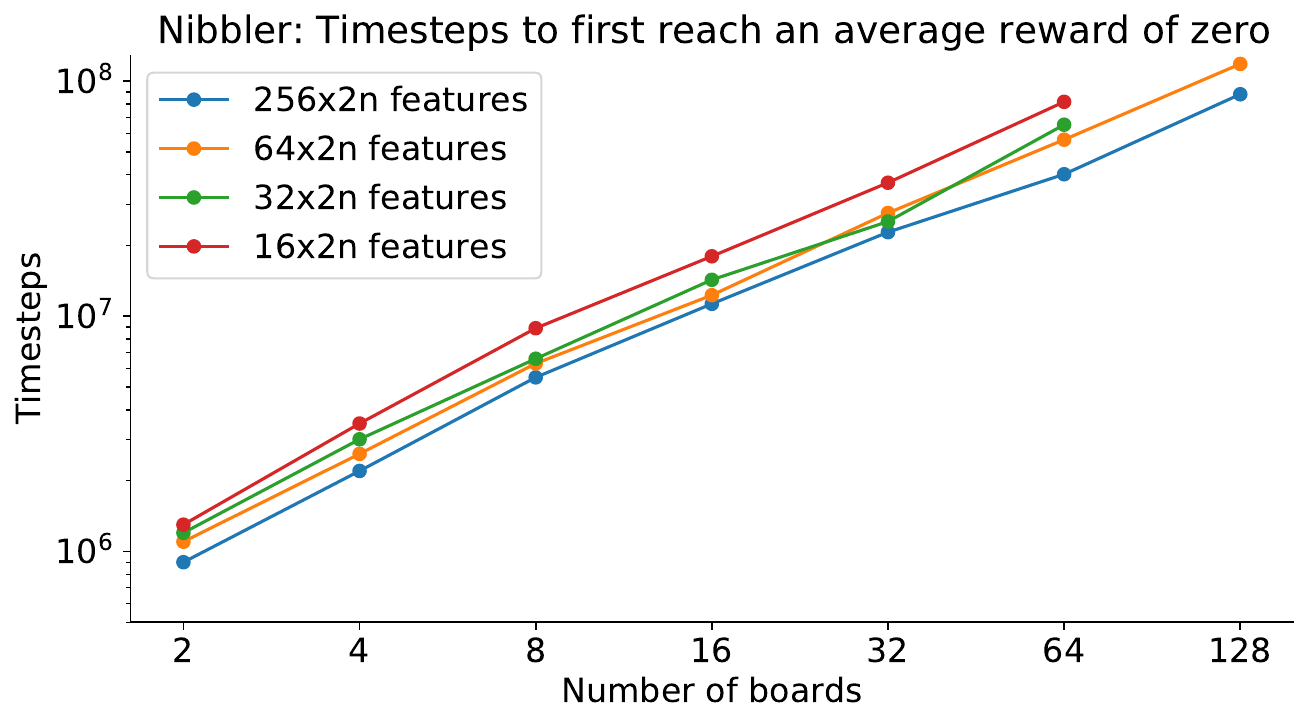}
\hfill
\includegraphics[height=1.5in]{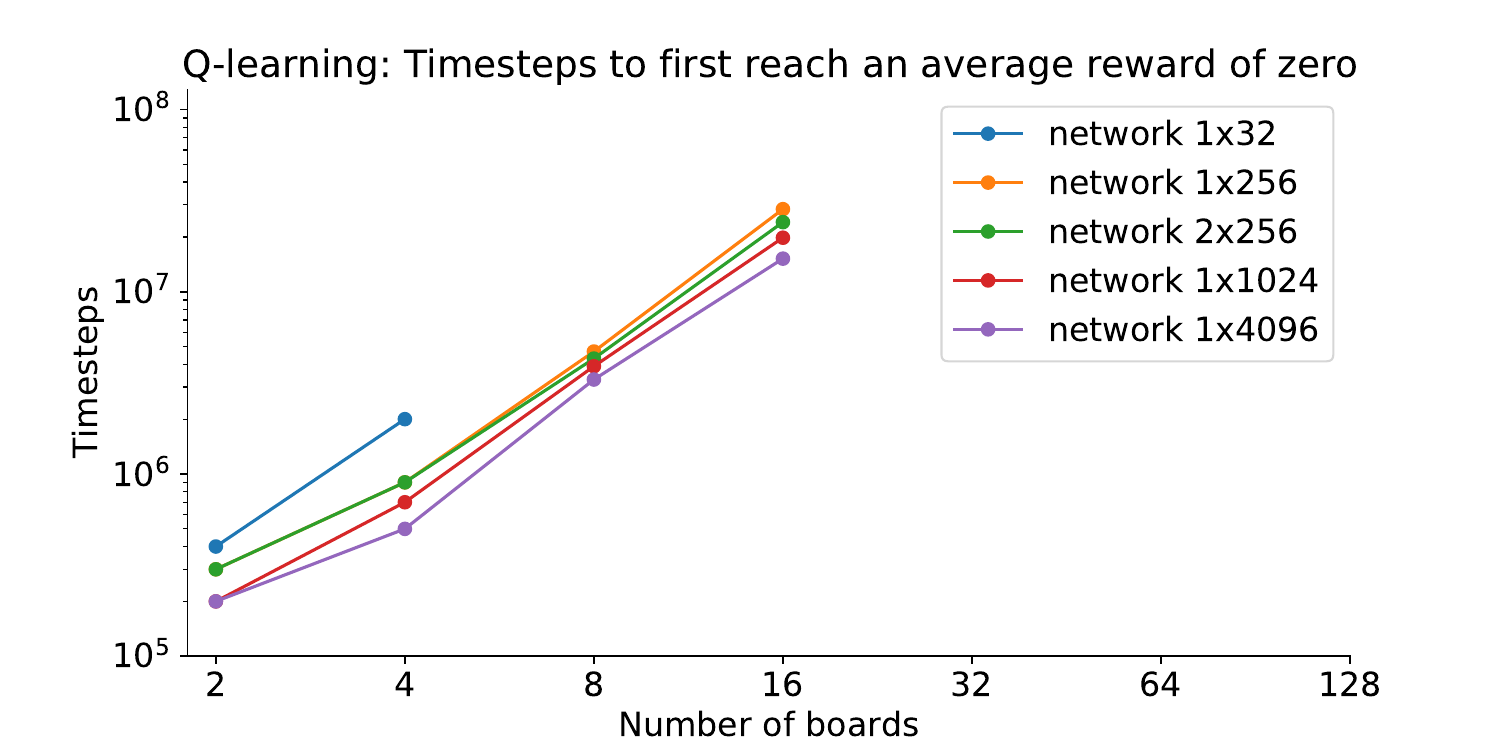}\\
\caption{Performance scales with available resources. We see that both the Nibbler architecture (left) and a fully connected deep Q-learning algorithm (right) reduce performance in proportion to the available hidden units, across problem scales in a predictable manner. However, the direct increase of resources in the Q-learning algorithm fails to improve the slope---indicating that the algorithm is scaling worse with the problem dimension.  Moreover, if we measure the number of network connections needed for backpropagation, a weak version of the Nibbler algorithm at 32 hidden features/question on 16 boards has $32 \times 16 \times 2=1024$ hidden features, and outperforms a Q-learning algorithm with 1024 hidden features in average reward and wall-clock time. Similar results hold for sweeps across learning rates.  On problems with this combinatorial form, the Nibbler algorithm demonstrates the benefits of adapting the network architecture in addition to the network parameters.  These results are averaged over 3 seeds per point.\label{fig:resource-scaling}}
\end{figure}

Our next experiment was to examine the effect on performance when we manipulate the size of the network but keep fixed step-sizes.  As shown in Figure~\ref{fig:resource-scaling}, both Nibbler and Q-learning show an improved time to threshold with increased number of hidden units (for the regime where they do not diverge).  This is a desirable qualitative property for an learning algorithm, namely increasing computational resources yields {\em faster} learning. 
However, adding weight resources to Nibbler or deep Q-learning algorithms (or reducing them) does not change the rate of improvement - the slopes of the zero crossing lines are not measurably changed in the regime where there is learning (compare with Figure~\ref{fig:nibble-scaling}).  \\

\subsection{Nibbler performance scales on domains with heterogeneous components}
\begin{figure}
\centerline{\includegraphics[height=2in]{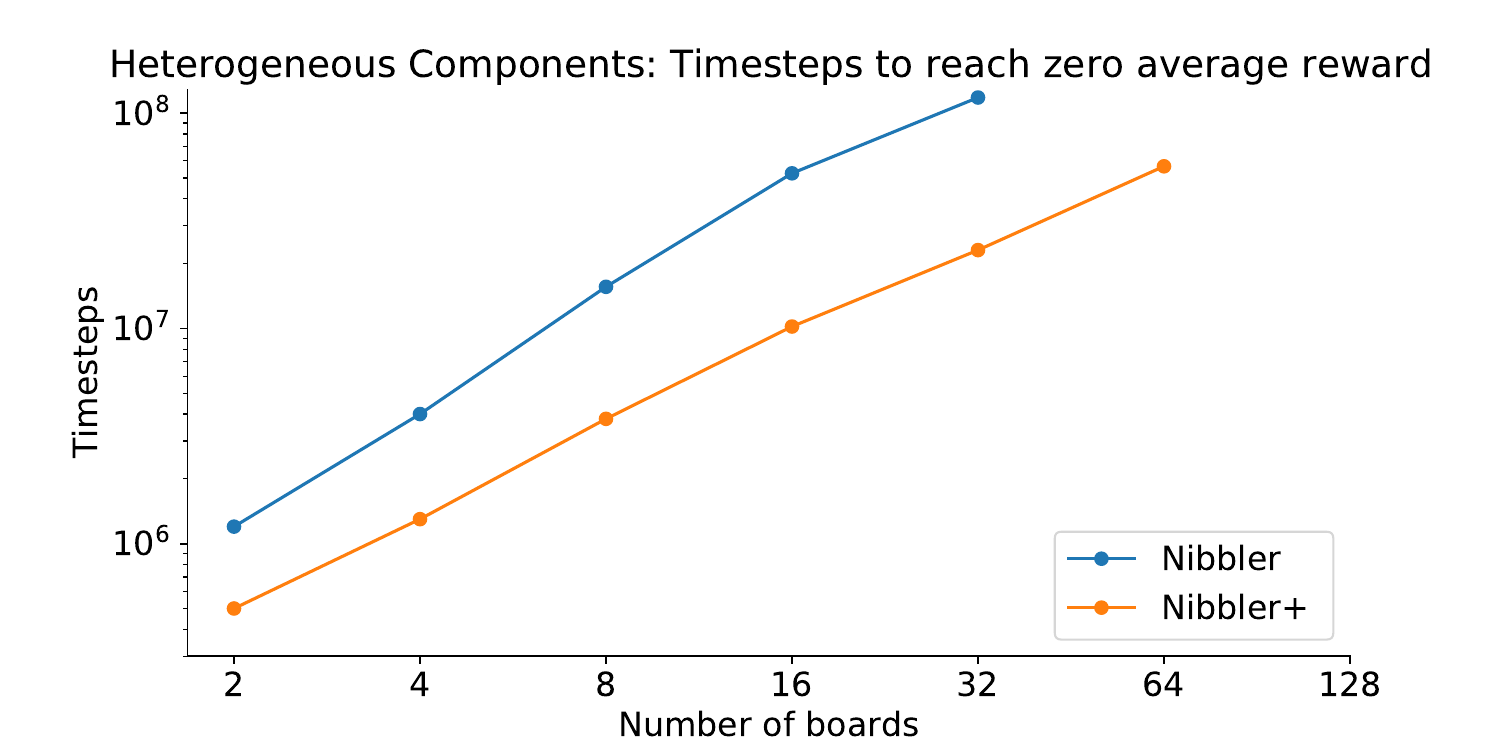}}
\caption{The Nibbler algorithm exhibits reliable scaling when the individual boards have varied sizes and dynamics.  The method shown as Nibbler+ is the same algorithm with the higher step-size choice of $\alpha=0.005\sqrt{2}/\sqrt{h}$, instead of $\alpha=0.001\sqrt{2}/\sqrt{h}$ used by default in Nibbler.  These results are averaged over 5 seeds.\label{fig:het}}
\end{figure}

 The Nibbler algorithm also behaves well when the individual components have substantially different internal dynamics.  We tested the algorithm on environments where the catch boards had varying numbers of rows (from 5 to 10), wind blowing the ball consistently left or right, and selecting the random variables of  $p_{arrival}$ and $p_{reward}$ from the uniform distribution over the interval $[0.05,1]$.  These choices alter the magnitude of the average reward for a random policy from the original problem, but they leave the attainment of zero average reward as an informative baseline. For these simple variations, the algorithm exhibits similar scaling properties, shown for two different choices of step-size in Figure~\ref{fig:het}.  A separate environment variation that caused both prediction and performance to become very noisy (a wind moves the ball stochastically) exhibited substantially worse performance and is not shown. \\

\section{Discussion}

The arguments thus far have pointed to a direction for studying how RL algorithms can scale. The primary goal has been to develop a regime for studying the phenomena, and to understand whether the phenomena of scaling could be instantiated in any computational regime.  The above results suggest they can be fruitfully studied.  The generality of the approach remains unclear from these experiments, as we have constructed the measures, the algorithms, and the domain to capture the primary facets of the phenomena of interest.   Hyper-parameter settings in the Nibbler algorithm and experiments are tuned to these constructed environments in ways that might be difficult to tune for non-constructed environments.  Further work is required to see whether this approach is useful for non-synthetic data (where data comes from the real world).  Additional work is likely required to make the Nibbler algorithm applicable across generic domains without hyperparameter tuning.  Further experimental runs with more seeds would provide more evidence for the differences between algorithms shown in the results. One potentially interesting note is that the problem space considered here for scalable and efficient learning relies on both non-episodic (continuing) domains and nonlinear function approximation. This is an under-explored problem regime within modern reinforcement learning research and it may have many additional research insights to offer. 

One insight from ML environments is that we can develop empirical scaling laws for algorithms.  This has been pursued extensively in the deep learning community, where scaling algorithms to large datasets has transformed language models~\autocite{gpt3_2020,dmgopher2021} with guidance for how to deploy data and compute resources~\autocite{kaplan2020,Hoffmann2022}.  Scaling within RL has seen substantial theoretical work presented within current formalisms~\autocite{szepesvari2010algorithms}, but many empirical concerns around domain scaling are dominated by the role of exploration~\autocite{Osband2019}. Some theoretical work~\autocite{abbe2021staircase} has examined the use of scalable function spaces structured into a ``staircase" motivated by the empirical success of ResNets, and a similar analysis may apply to the structured function space of the Nibbler algorithm.    Although substantial efforts have been made to increase the generality of environments RL algorithms are tested on~\autocite{machado2018revisiting}, little work directly addresses scaling with non-linear function approximation.  

On the algorithmic side, some recent works have pushed towards handling large unstructured data.  The feature selection mechanism for GVF answers in Nibbler is extended from previous work in the pure prediction setting~\autocite{martin2021}.  That earlier algorithm did not refine the non-learner features, and the GVF questions were selected at random instead of being informed by the reward signal.   Another line of work for finding useful nonlinear features in unstructured data has been in online supervised learning~\autocite{sutton1993online,mahmood2013representation}.   Work on generalizing beyond convolutions has shown performant (though computationally expensive) supervised learning using regularization with n a massive parameter approximation space~\autocite{neyshabur2020towards,molchanov2019pruning} and also with vision transformers~\autocite{dosovitskiy2020image}.  Even more computationally expensive mechanisms are available that search through a space of architectures seeking a performant hypothesis space, either in the architecture directly with hard constraints~\autocite{elsken2019neural}, or in the network initialization weights as a soft constraint~\autocite{frankle2018lottery,evci2020rigging,finn2017model}.   A more closely related approach of directly removing irrelevant connections for deep RL algorithms has been proposed recently~\autocite{sokar2022dynamic,grooten2023automatic}, independent of this work.  Those works rely on removing small weight connections, but do not require the explicit formation of GVF subproblems, and it would be interesting point of comparison. 

Several lines of research have used reward prediction as a mechanism to support additional predictive structure.  An early example is in planning, where a linear reward model is a central piece to the Linear Dyna algorithm~\autocite{sutton2008dyna,parr2008analysis}, and related work on planning with option models~\autocite{Sorg2010,Yao2014universal,sutton2022rewardrespecting}.  A related line of work is on successor features~\autocite{barreto2017successor} which also uses a reward model and predictions of the features, but not with the computational benefits of representing internal structure.  A more partitioned approach used handcrafted reward features (and function spaces) for the specific Atari game of Ms.\ Pacman~\autocite{van2017hybrid}.  

The role of using a collection of auxiliary tasks (including GVFs) to shape the representation in a deep neural net, was pursued in the UNREAL architecture~\autocite{jaderberg2016reinforcement}.  Other approaches to the collection of auxiliary tasks pursue them as a set of interrelated control problems~\autocite{riedmiller18a}, most commonly examining transfer across a set of episodic environments that can form an implicit curriculum~\autocite{heess2017emergence}.

Finally the pursuit of strictly incremental deep RL algorithms is not unique to this work or completed by it.  Much like the major transition in supervised learning from batch to minibatch methods has been changed by the size of the datastreams people consider~\autocite{bottou2007tradeoffs},
 the boundaries between fixed-dataset RL methods (approximate dynamic programming), online batch RL methods~\autocite{mnih2015human,riedmiller2005neural}, and strictly incremental RL methods is a fluid one.  Earlier examples of successful incremental training of a deep network during policy improvement exist~\autocite{mnih2016asynchronous,van_Hasselt_aaai_2021}. 
 Although the experiments shown in this paper demonstrate robust results, the algorithms have only been tested on a few synthetic environments with a fast mixing time.  More research will be required to evaluate and improve the robustness of these function spaces and step-size optimizers on a greater diversity of problems in this strictly incremental setting. Potentially even more changes will be required for continual learning problems.

\section{Conclusions}

We have provided a mechanism for defining families of scalable RL environments with unstructured observations, along with a metric for evaluating the growth in sample complexity across problem scales.  We have provided a novel deep RL algorithm that combines linear and deep learning for reinforcement learning to achieve a dynamic neural network architecture: it connects an online experience stream with unknown structure observation structure to a fixed deep RL network architecture guided by statistics acquired by the linear learner.  The proposed method substantially outperforms conventional deep RL algorithms using fixed network architectures on a family of synthetic problems.  The algorithm  exhibits favorable and predictable performance when the algorithm's computational resources are scaled along with problem's state space, and that standard fixed architecture deep RL solutions do not perform as well at larger scales across a range of hyperparameter settings.  We demonstrated a range of ablations showing the contributions of each of the core ideas.

\section{Acknowledgements}
Many individuals have contributed towards this research with their insights and conversations.  The authors gratefully acknowledge the comments from Thomas Degris, Hado van Hasselt,  Michael Bowling, Rich Sutton, Tyler Jackson, and John Martin.  The errors of the paper belong entirely to the authors.
\printbibliography

@book{Eagleman2020,
  title={Livewired: The inside story of the ever-changing brain},
  author={Eagleman, David},
  year={2020},
  publisher={Canongate Books}
}

@misc{abbe2021staircase,
      title={The staircase property: How hierarchical structure can guide deep learning}, 
      author={Emmanuel Abbe and Enric Boix-Adsera and Matthew Brennan and Guy Bresler and Dheeraj Nagaraj},
      year={2021},
      eprint={2108.10573},
      archivePrefix={arXiv},
      primaryClass={cs.LG}
}

@article{singh1997dynamically,
  title={How to dynamically merge {M}arkov decision processes},
  author={Singh, Satinder and Cohn, David},
  journal={Advances in neural information processing systems},
  volume={10},
  year={1997}
}

@inproceedings{koller1999computing,
  title={Computing factored value functions for policies in structured MDPs},
  author={Koller, Daphne and Parr, Ronald},
  booktitle={IJCAI},
  volume={99},
  pages={1332--1339},
  year={1999}
}

@article{boutilier2000stochastic,
  title={Stochastic dynamic programming with factored representations},
  author={Boutilier, Craig and Dearden, Richard and Goldszmidt, Mois{\'e}s},
  journal={Artificial intelligence},
  volume={121},
  number={1-2},
  pages={49--107},
  year={2000},
  publisher={Elsevier}
}

@book{pavlov-quote,
title={Conditioned Reflexes},
author={I.P. Pavlov},
year={2015},
pages={14},
publisher={Martino Publishing},
note={originally published in 1927}, 
address={Mansfield Centre, CT}
}

@book{Doidge, address={Harlow, England}, title={The brain that changes itself: Stories of personal triumph from the frontiers of brain science}, ISBN={9780141038872}, publisher={Penguin Books}, author={Doidge, Norman}, year={2008}, language={en} }

@misc{mitchell1997machine,
  title={Machine learning},
  author={Mitchell, Tom M and others},
  year={1997},
  publisher={McGraw-hill New York}
}

@book{mackay2003information,
  title={Information theory, inference and learning algorithms},
  author={MacKay, David JC},
  year={2003},
  publisher={Cambridge university press}
}

@article{machado2018revisiting,
  title={Revisiting the arcade learning environment: Evaluation protocols and open problems for general agents},
  author={Machado, Marlos and Bellemare, Marc G and Talvitie, Erik and Veness, Joel and Hausknecht, Matthew and Bowling, Michael},
  journal={Journal of Artificial Intelligence Research},
  volume={61},
  pages={523--562},
  year={2018}
}

@inproceedings{Sutton96generalizationin,
    author = {Sutton, Richard},
    title = {Generalization in Reinforcement Learning: Successful Examples Using Sparse Coarse Coding},
    booktitle = {Advances in Neural Information Processing Systems 8},
    year = {1996},
    pages = {1038--1044},
    publisher = {MIT Press}
}

@article{tassa2018deepmind,
  title={Deepmind control suite},
  author={Tassa, Yuval and Doron, Yotam and Muldal, Alistair and Erez, Tom and Li, Yazhe and Casas, Diego de Las and Budden, David and Abdolmaleki, Abbas and Merel, Josh and Lefrancq, Andrew and others},
  journal={arXiv preprint arXiv:1801.00690},
  year={2018}
}

@article{brockman2016openai,
  title={Openai gym},
  author={Brockman, Greg and Cheung, Vicki and Pettersson, Ludwig and Schneider, Jonas and Schulman, John and Tang, Jie and Zaremba, Wojciech},
  journal={arXiv preprint arXiv:1606.01540},
  year={2016}
}

@misc{Osband2019,
  doi = {10.48550/ARXIV.1908.03568},
  % url = {https://arxiv.org/abs/1908.03568},
  author = {Osband, Ian and Doron, Yotam and Hessel, Matteo and Aslanides, John and Sezener, Eren and Saraiva, Andre and McKinney, Katrina and Lattimore, Tor and Szepesvari, Csaba and Singh, Satinder and Van Roy, Benjamin and Sutton, Richard and Silver, David and {v}an Hasselt, Hado},
  keywords = {Machine Learning (cs.LG), Artificial Intelligence (cs.AI), Machine Learning (stat.ML), FOS: Computer and information sciences, FOS: Computer and information sciences},
  title = {Behaviour Suite for Reinforcement Learning},
  publisher = {arXiv},
  year = {2019},
  copyright = {arXiv.org perpetual, non-exclusive license}
}

@article{bhatnagar2009natural,
  title={Natural actor--critic algorithms},
  author={Bhatnagar, Shalabh and Sutton, Richard and Ghavamzadeh, Mohammad and Lee, Mark},
  journal={Automatica},
  volume={45},
  number={11},
  pages={2471--2482},
  year={2009},
  publisher={Elsevier}
}

@article{elsken2019neural,
  title={Neural architecture search: A survey},
  author={Elsken, Thomas and Metzen, Jan Hendrik and Hutter, Frank},
  journal={The Journal of Machine Learning Research},
  volume={20},
  number={1},
  pages={1997--2017},
  year={2019},
  publisher={JMLR. org}
}

@article{zoph2016neural,
  title={Neural architecture search with reinforcement learning},
  author={Zoph, Barret and Le, Quoc V},
  journal={arXiv preprint arXiv:1611.01578},
  year={2016}
}

@inproceedings{finn2017model,
  title={Model-agnostic meta-learning for fast adaptation of deep networks},
  author={Finn, Chelsea and Abbeel, Pieter and Levine, Sergey},
  booktitle={International conference on machine learning},
  pages={1126--1135},
  year={2017},
  organization={PMLR}
}

@inproceedings{frankle2018lottery,
  title={The Lottery Ticket Hypothesis: Finding Sparse, Trainable Neural Networks},
  author={Frankle, Jonathan and Carbin, Michael},
  booktitle={International Conference on Learning Representations},
  year={2018}
}

@inproceedings{evci2020rigging,
  title={Rigging the lottery: Making all tickets winners},
  author={Evci, Utku and Gale, Trevor and Menick, Jacob and Castro, Pablo Samuel and Elsen, Erich},
  booktitle={International Conference on Machine Learning},
  pages={2943--2952},
  year={2020},
  organization={PMLR}
}

@article{sutton1993online,
  title={Online learning with random representations},
  author={Sutton, Richard and Whitehead, Steven D},
  booktitle={Proceedings of the Tenth International Conference on Machine Learning},
  pages={314--321},
  year={1993}
}

@book{sutton2018reinforcement,
  title={Reinforcement learning: An introduction},
  author={Sutton, Richard and Barto, Andrew},
  year={2018},
  publisher={MIT press}
}

@article{littman2001predictive,
  title={Predictive representations of state},
  author={Littman, Michael and Sutton, Richard and Singh, Satinder},
  journal={Advances in neural information processing systems},
  volume={14},
  year={2001}
}

@article{boots2011closing,
  title={Closing the learning-planning loop with predictive state representations},
  author={Boots, Byron and Siddiqi, Sajid M and Gordon, Geoffrey J},
  journal={The International Journal of Robotics Research},
  volume={30},
  number={7},
  pages={954--966},
  year={2011},
  publisher={SAGE Publications Sage UK: London, England}
}

@article{modayil2014multi,
  title={Multi-timescale nexting in a reinforcement learning robot},
  author={Modayil, Joseph and White, Adam and Sutton, Richard},
  journal={Adaptive Behavior},
  volume={22},
  number={2},
  pages={146--160},
  year={2014},
  publisher={SAGE Publications Sage UK: London, England}
}

@article{jaderberg2016reinforcement,
  title={Reinforcement learning with unsupervised auxiliary tasks},
  author={Jaderberg, Max and Mnih, Volodymyr and Czarnecki, Wojciech Marian and Schaul, Tom and Leibo, Joel Z and Silver, David and Kavukcuoglu, Koray},
  journal={arXiv preprint arXiv:1611.05397},
  year={2016}
}

@inproceedings{parr2008analysis,
  title={An analysis of linear models, linear value-function approximation, and feature selection for reinforcement learning},
  author={Parr, Ronald and Li, Lihong and Taylor, Gavin and Painter-Wakefield, Christopher and Littman, Michael L},
  booktitle={Proceedings of the 25th international conference on Machine learning},
  pages={752--759},
  year={2008}
}

@article{sutton2008dyna,
  title={Dyna-style planning with linear function approximation and prioritized sweeping},
  author={Sutton, Richard and Szepesv{\'a}ri, Csaba and Geramifard, Alborz and Bowling, Michael P},
  journal={Proceedings of the Twenty-Fourth Conference on Uncertainty in Artificial Intelligence (UAI2008)},
  year={2008}
}

@article{dayan1993improving,
  title={Improving generalization for temporal difference learning: The successor representation},
  author={Dayan, Peter},
  journal={Neural Computation},
  volume={5},
  number={4},
  pages={613--624},
  year={1993},
  publisher={MIT Press}
}

@article{barreto2017successor,
  title={Successor features for transfer in reinforcement learning},
  author={Barreto, Andr{\'e} and Dabney, Will and Munos, R{\'e}mi and Hunt, Jonathan J and Schaul, Tom and van Hasselt, Hado and Silver, David},
  journal={Advances in neural information processing systems},
  volume={30},
  year={2017}
}

@misc{kaplan2020,
  doi = {10.48550/ARXIV.2001.08361},
  
  % url = {https://arxiv.org/abs/2001.08361},
  
  author = {Kaplan, Jared and McCandlish, Sam and Henighan, Tom and Brown, Tom B. and Chess, Benjamin and Child, Rewon and Gray, Scott and Radford, Alec and Wu, Jeffrey and Amodei, Dario},
  
  keywords = {Machine Learning (cs.LG), Machine Learning (stat.ML), FOS: Computer and information sciences, FOS: Computer and information sciences},
  
  title = {Scaling Laws for Neural Language Models},
  
  publisher = {arXiv},
  
  year = {2020},
  
  copyright = {arXiv.org perpetual, non-exclusive license}
}

@misc{Hoffmann2022,
  doi = {10.48550/ARXIV.2203.15556},
  
 %  url = {https://arxiv.org/abs/2203.15556},
  
  author = {Hoffmann, Jordan and Borgeaud, Sebastian and Mensch, Arthur and Buchatskaya, Elena and Cai, Trevor and Rutherford, Eliza and Casas, Diego de Las and Hendricks, Lisa Anne and Welbl, Johannes and Clark, Aidan and Hennigan, Tom and Noland, Eric and Millican, Katie and Driessche, George van den and Damoc, Bogdan and Guy, Aurelia and Osindero, Simon and Simonyan, Karen and Elsen, Erich and Rae, Jack W. and Vinyals, Oriol and Sifre, Laurent},
  
  keywords = {Computation and Language (cs.CL), Machine Learning (cs.LG), FOS: Computer and information sciences, FOS: Computer and information sciences},
  
  title = {Training Compute-Optimal Large Language Models},
  
  publisher = {arXiv},
  
  year = {2022},
  
  copyright = {arXiv.org perpetual, non-exclusive license}
}

@misc{dmgopher2021,
  doi = {10.48550/ARXIV.2112.11446},
  
  % url = {https://arxiv.org/abs/2112.11446},
  
  author = {Rae, Jack W. and Borgeaud, Sebastian and Cai, Trevor and Millican, Katie and Hoffmann, Jordan and Song, Francis and Aslanides, John and Henderson, Sarah and Ring, Roman and Young, Susannah and Rutherford, Eliza and Hennigan, Tom and Menick, Jacob and Cassirer, Albin and Powell, Richard and Driessche, George van den and Hendricks, Lisa Anne and Rauh, Maribeth and Huang, Po-Sen and Glaese, Amelia and Welbl, Johannes and Dathathri, Sumanth and Huang, Saffron and Uesato, Jonathan and Mellor, John and Higgins, Irina and Creswell, Antonia and McAleese, Nat and Wu, Amy and Elsen, Erich and Jayakumar, Siddhant and Buchatskaya, Elena and Budden, David and Sutherland, Esme and Simonyan, Karen and Paganini, Michela and Sifre, Laurent and Martens, Lena and Li, Xiang Lorraine and Kuncoro, Adhiguna and Nematzadeh, Aida and Gribovskaya, Elena and Donato, Domenic and Lazaridou, Angeliki and Mensch, Arthur and Lespiau, Jean-Baptiste and Tsimpoukelli, Maria and Grigorev, Nikolai and Fritz, Doug and Sottiaux, Thibault and Pajarskas, Mantas and Pohlen, Toby and Gong, Zhitao and Toyama, Daniel and d'Autume, Cyprien de Masson and Li, Yujia and Terzi, Tayfun and Mikulik, Vladimir and Babuschkin, Igor and Clark, Aidan and Casas, Diego de Las and Guy, Aurelia and Jones, Chris and Bradbury, James and Johnson, Matthew and Hechtman, Blake and Weidinger, Laura and Gabriel, Iason and Isaac, William and Lockhart, Ed and Osindero, Simon and Rimell, Laura and Dyer, Chris and Vinyals, Oriol and Ayoub, Kareem and Stanway, Jeff and Bennett, Lorrayne and Hassabis, Demis and Kavukcuoglu, Koray and Irving, Geoffrey},
  
  keywords = {Computation and Language (cs.CL), Artificial Intelligence (cs.AI), FOS: Computer and information sciences, FOS: Computer and information sciences},
  
  title = {Scaling Language Models: Methods, Analysis \& Insights from Training Gopher},
  
  publisher = {arXiv},
  
  year = {2021},
  
  copyright = {arXiv.org perpetual, non-exclusive license}
}

@inproceedings{gpt3_2020,
 author = {Brown, Tom and Mann, Benjamin and Ryder, Nick and Subbiah, Melanie and Kaplan, Jared D and Dhariwal, Prafulla and Neelakantan, Arvind and Shyam, Pranav and Sastry, Girish and Askell, Amanda and Agarwal, Sandhini and Herbert-Voss, Ariel and Krueger, Gretchen and Henighan, Tom and Child, Rewon and Ramesh, Aditya and Ziegler, Daniel and Wu, Jeffrey and Winter, Clemens and Hesse, Chris and Chen, Mark and Sigler, Eric and Litwin, Mateusz and Gray, Scott and Chess, Benjamin and Clark, Jack and Berner, Christopher and McCandlish, Sam and Radford, Alec and Sutskever, Ilya and Amodei, Dario},
 booktitle = {Advances in Neural Information Processing Systems},
 editor = {H. Larochelle and M. Ranzato and R. Hadsell and M.F. Balcan and H. Lin},
 pages = {1877--1901},
 publisher = {Curran Associates, Inc.},
 title = {Language Models are Few-Shot Learners},
 %url = {https://proceedings.neurips.cc/paper/2020/file/1457c0d6bfcb4967418bfb8ac142f64a-Paper.pdf},
 volume = {33},
 year = {2020}
}

@misc{martin2021,
  doi = {10.48550/ARXIV.2106.09776},
  
  % url = {https://arxiv.org/abs/2106.09776},
  
  author = {Martin, John D. and Modayil, Joseph},
  
  keywords = {Machine Learning (cs.LG), Artificial Intelligence (cs.AI), FOS: Computer and information sciences, FOS: Computer and information sciences},
  
  title = {Adapting the Function Approximation Architecture in Online Reinforcement Learning},
  
  publisher = {arXiv},
  
  year = {2021},
  
  copyright = {arXiv.org perpetual, non-exclusive license}
}

@inproceedings{mahmood2013representation,
  title={Representation Search through Generate and Test.},
  author={Mahmood, Ashique Rupam and Sutton, Richard},
  year={2013}
}

@article{neyshabur2020towards,
  title={Towards learning convolutions from scratch},
  author={Neyshabur, Behnam},
  journal={Advances in Neural Information Processing Systems},
  volume={33},
  year={2020}
}

@inproceedings{molchanov2019pruning,
  title={Pruning convolutional neural networks for resource efficient inference},
  author={Molchanov, P and Tyree, S and Karras, T and Aila, T and Kautz, J},
  booktitle={5th International Conference on Learning Representations, ICLR 2017-Conference Track Proceedings},
  year={2019}
}

@article{dosovitskiy2020image,
  title={An image is worth 16x16 words: Transformers for image recognition at scale},
  author={Dosovitskiy, Alexey and Beyer, Lucas and Kolesnikov, Alexander and Weissenborn, Dirk and Zhai, Xiaohua and Unterthiner, Thomas and Dehghani, Mostafa and Minderer, Matthias and Heigold, Georg and Gelly, Sylvain and others},
  journal={arXiv preprint arXiv:2010.11929},
  year={2020}
}

@inproceedings{Sorg2010,
author = {Sorg, Jonathan and Singh, Satinder},
title = {Linear Options},
year = {2010},
isbn = {9780982657119},
publisher = {International Foundation for Autonomous Agents and Multiagent Systems},
address = {Richland, SC},
booktitle = {Proceedings of the 9th International Conference on Autonomous Agents and Multiagent Systems: Volume 1 - Volume 1},
pages = {31–38},
numpages = {8},
location = {Toronto, Canada},
series = {AAMAS '10}
}

@article{Yao2014universal,
  title={Universal option models},
  author={Yao, Hengshaui and Szepesvari, Csaba and Sutton, Richard and Modayil, Joseph and Bhatnagar, Shalabh and others},
  journal={Advances in Neural Information Processing Systems},
  volume={27},
  year={2014}
}

@book{szepesvari2010algorithms,
  title={Algorithms for reinforcement learning},
  author={Szepesv{\'a}ri, Csaba},
  series={Synthesis lectures on artificial intelligence and machine learning},
  volume={4},
  number={1},
  year={2010},
  publisher={Morgan \& Claypool Publishers}
}

@InProceedings{riedmiller18a,
  title = 	 {Learning by Playing Solving Sparse Reward Tasks from Scratch},
  author =       {Riedmiller, Martin and Hafner, Roland and Lampe, Thomas and Neunert, Michael and Degrave, Jonas and van de Wiele, Tom and Mnih, Vlad and Heess, Nicolas and Springenberg, Jost Tobias},
  booktitle = 	 {Proceedings of the 35th International Conference on Machine Learning},
  pages = 	 {4344--4353},
  year = 	 {2018},
  editor = 	 {Dy, Jennifer and Krause, Andreas},
  volume = 	 {80},
  series = 	 {Proceedings of Machine Learning Research},
  month = 	 Jul,
  publisher =    {PMLR},
  url = 	 {https://proceedings.mlr.press/v80/riedmiller18a.html}
}

@article{heess2017emergence,
  title={Emergence of locomotion behaviours in rich environments},
  author={Heess, Nicolas and TB, Dhruva and Sriram, Srinivasan and Lemmon, Jay and Merel, Josh and Wayne, Greg and Tassa, Yuval and Erez, Tom and Wang, Ziyu and Eslami, SM and others},
  journal={arXiv preprint arXiv:1707.02286},
  year={2017}
}

@article{bottou2007tradeoffs,
  title={The tradeoffs of large scale learning},
  author={Bottou, L{\'e}on and Bousquet, Olivier},
  journal={Advances in neural information processing systems},
  volume={20},
  year={2007}
}

@inproceedings{mnih2016asynchronous,
  title={Asynchronous methods for deep reinforcement learning},
  author={Mnih, Volodymyr and Badia, Adria Puigdomenech and Mirza, Mehdi and Graves, Alex and Lillicrap, Timothy and Harley, Tim and Silver, David and Kavukcuoglu, Koray},
  booktitle={International conference on machine learning},
  pages={1928--1937},
  year={2016},
  organization={PMLR}
}

@article{van2017hybrid,
  title={Hybrid reward architecture for reinforcement learning},
  author={{v}an Seijen, Harm and Fatemi, Mehdi and Romoff, Joshua and Laroche, Romain and Barnes, Tavian and Tsang, Jeffrey},
  sortname={Seijen, Harm van},
  journal={Advances in Neural Information Processing Systems},
  volume={30},
  year={2017}
}

@misc{bitterLesson,
  title = {The Bitter Lesson},
  author = {Sutton, Richard},
  year = {2019},
  howpublished = {\url{http://www.incompleteideas.net/IncIdeas/BitterLesson.html}},
  note = {Accessed: 2022-06-27}
}

@article{Schlegel_2021,
	doi = {10.1613/jair.1.12105},
  
	% url = {https://doi.org/10.1613%2Fjair.1.12105},
  
	year = 2021,
	month = jan,
  
	publisher = {{AI} Access Foundation},
  
	volume = {70},
  
	pages = {497--543},
  
	author = {Matthew Schlegel and Andrew Jacobsen and Zaheer Abbas and Andrew Patterson and Adam White and Martha White},
  
	title = {General Value Function Networks},
  
	journal = {Journal of Artificial Intelligence Research}
}

@misc{sutton2022rewardrespecting,
      title={Reward-Respecting Subtasks for Model-Based Reinforcement Learning}, 
      author={Richard Sutton and Marlos Machado and G. Zacharias Holland and David Szepesvari and Finbarr Timbers and Brian Tanner and Adam White},
      year={2022},
      eprint={2202.03466},
      archivePrefix={arXiv},
      primaryClass={cs.LG}
}

@inproceedings{riedmiller2005neural,
  title={Neural fitted Q iteration--first experiences with a data efficient neural reinforcement learning method},
  author={Riedmiller, Martin},
  booktitle={European conference on machine learning},
  pages={317--328},
  year={2005},
  organization={Springer}
}

@article{mnih2015human,
  title={Human-level control through deep reinforcement learning},
  author={Mnih, Volodymyr and Kavukcuoglu, Koray and Silver, David and Rusu, Andrei A and Veness, Joel and Bellemare, Marc G and Graves, Alex and Riedmiller, Martin and Fidjeland, Andreas K and Ostrovski, Georg and others},
  journal={Nature},
  volume={518},
  number={7540},
  pages={529--533},
  year={2015},
  publisher={Nature Publishing Group}
}

@article{van_Hasselt_aaai_2021, 
title={Expected Eligibility Traces}, volume={35}, url={https://ojs.aaai.org/index.php/AAAI/article/view/17200}, 
 number={11}, 
 journal={Proceedings of the AAAI Conference on Artificial Intelligence}, 
 author={{v}an Hasselt, Hado and Madjiheurem, Sephora and Hessel, Matteo and Silver, David and Barreto, André and Borsa, Diana},
 sortname={Hasselt, Hado van and Madjiheurem, Sephora},
 year={2021}, month=May, 
 pages={9997-10005} }

@inproceedings{Zheng2021LearningSR,
  title={Learning State Representations from Random Deep Action-conditional Predictions},
  author={Zeyu Zheng and Vivek Veeriah and Risto Vuorio and Richard L. Lewis and Satinder Singh},
  booktitle={Advances in Neural Information Processing Systems 34},
  year={2021}
}

@inproceedings{sutton2011horde,
  title={Horde: A scalable real-time architecture for learning knowledge from unsupervised sensorimotor interaction},
  author={Sutton, Richard and Modayil, Joseph and Delp, Michael and Degris, Thomas and Pilarski, Patrick M and White, Adam and Precup, Doina},
  booktitle={The 10th International Conference on Autonomous Agents and Multiagent Systems-Volume 2},
  pages={761--768},
  year={2011}
}

@inproceedings{wiering2009qv,
  title={The QV family compared to other reinforcement learning algorithms},
  author={Wiering, Marco A and {v}an Hasselt, Hado},
  booktitle={2009 IEEE Symposium on Adaptive Dynamic Programming and Reinforcement Learning},
  pages={101--108},
  year={2009},
  organization={IEEE}
}

@inproceedings{hessel2018rainbow,
  title={Rainbow: Combining improvements in deep reinforcement learning},
  author={Hessel, Matteo and Modayil, Joseph and {v}an Hasselt, Hado and Schaul, Tom and Ostrovski, Georg and Dabney, Will and Horgan, Dan and Piot, Bilal and Azar, Mohammad and Silver, David},
  booktitle={Thirty-second AAAI conference on artificial intelligence},
  year={2018}
}

@book{dehaene2020,
  title={How we learn: Why brains learn better than any machine... for now},
  author={Dehaene, Stanislas},
  year={2020},
  publisher={Penguin}
}

@misc{grooten2023automatic,
      title={Automatic Noise Filtering with Dynamic Sparse Training in Deep Reinforcement Learning}, 
      author={Bram Grooten and Ghada Sokar and Shibhansh Dohare and Elena Mocanu and Matthew E. Taylor and Mykola Pechenizkiy and Decebal Constantin Mocanu},
      year={2023},
      eprint={2302.06548},
      archivePrefix={arXiv},
      primaryClass={cs.LG}
}

@misc{sokar2022dynamic,
      title={Dynamic Sparse Training for Deep Reinforcement Learning}, 
      author={Ghada Sokar and Elena Mocanu and Decebal Constantin Mocanu and Mykola Pechenizkiy and Peter Stone},
      year={2022},
      eprint={2106.04217},
      archivePrefix={arXiv},
      primaryClass={cs.LG}
}

@software{dqnzoo2020github,
  title = {{DQN} {Zoo}: Reference implementations of {DQN}-based agents},
  author = {John Quan and Georg Ostrovski},
  url = {http://github.com/deepmind/dqn_zoo},
  version = {1.2.0},
  year = {2020}
  }

\appendix

\section{More algorithm detail}
\begin{algorithm}
\caption{Incremental-top-k}
\begin{algorithmic}[1]
\Require $A_L \in \{1,...,m \}^k$ a list of length $k$ of the previously selected base feature indices in $\{1,\ldots,m\}$
\Require $A_M \in \{0,1 \}^m$ a list of length $m$, with $h$ ones indicating previously selected base feature indices 
\Require $\tau$ the threshold for swapping
\Require $U$, a vector holding the utility for each feature
\Ensure The most useful previously unselected  feature is swapped with the least useful previously selected feature, if the unselected feature's utility exceeds the selected feature's utility by a threshold $\tau$.
\State $low \leftarrow \arg\min_{i \in A_M} U[i]$
\State $high \leftarrow \arg\max_{i \not \in A_M} U[i]$
\State $pos \leftarrow \emptyset$
\State $changed \leftarrow U[low] + \tau < U[high]$
 \If {$ changed$ }
\State $A_M[low] \leftarrow 0$
\State $A_M[high] \leftarrow 1$
\State $pos \leftarrow \text{get-index}(low, A_L)$
\State $A_L[pos] \leftarrow high$
\EndIf
\State \Return $A_L, A_M, changed, pos$
\end{algorithmic}
\end{algorithm}

\begin{algorithm}
\caption{Co-opt SGD with momentum}
\begin{algorithmic}[1]
\Require $w$ the weights of interest
\Require $\vartheta$ is the momentum state
\Require $dx$ the gradient of loss with respect to the weights of interest
\Require $(\alpha,\nu) $ the stepsize and momentum hyperparameters
\State $\vartheta' \leftarrow \nu \vartheta +(1-\nu) dx$
\State $w' \leftarrow w + \alpha \vartheta'$
\State \Return $w', \vartheta'$
\end{algorithmic}
\end{algorithm}

\section{Additional sweeps and ablations}
\subsection{Fixed hyperparameter Nibbler algorithm on varying problem sizes }

\begin{figure}
    \centering
    \begin{tabular}{cc}
    \includegraphics[width=4in]{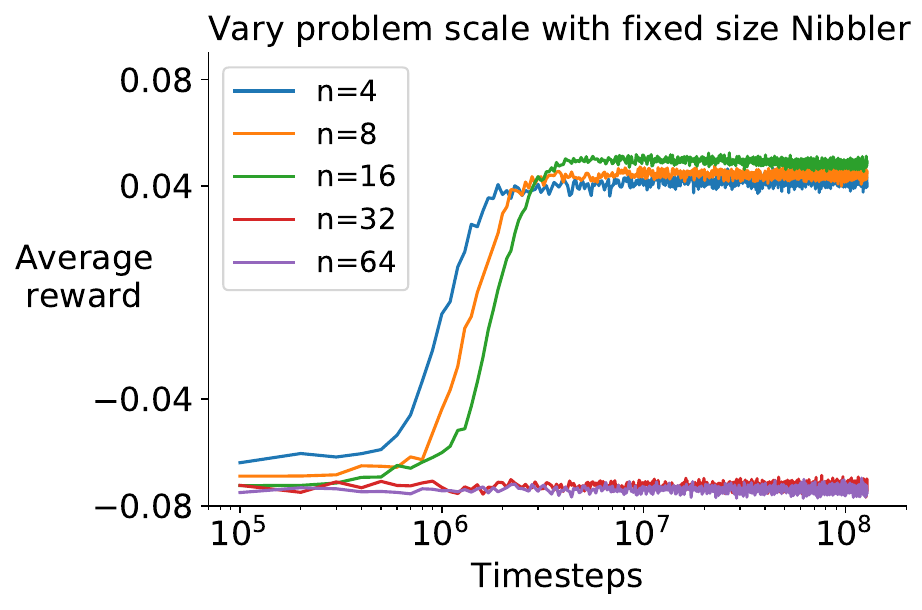} &
    \end{tabular}
    \caption{Examining mismatches between the problem scale and algorithm scale. Here, a nibbler algorithm with a fixed number of GVF questions is applied to coupled catch problems of different scales.  The algorithm performs well when adequately resourced for the problem, but performance collapses when adequate resources are not available.}
    \label{fig:vary-boards}
\end{figure}

We see the performance of the Nibbler algorithm as the problem scale is varied in Figure~\ref{fig:vary-boards}.  When inadequate resources are available, the performance of Nibbler collapses with 32 boards (as does the performance of Q-learning with 32 boards in Figure~\ref{fig:nibble-scaling}).  The complete performance collapse suggests running with a higher threshold than zero for $\tau$ could support a more graceful degradation in performance.

\begin{figure}
    \centering
    \includegraphics[height=2in]{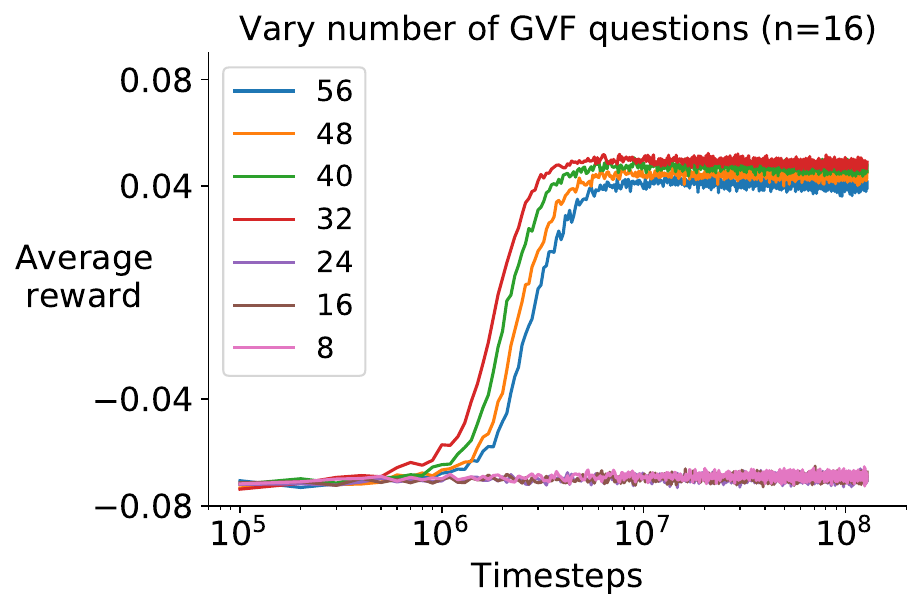}
    \includegraphics[height=2in]{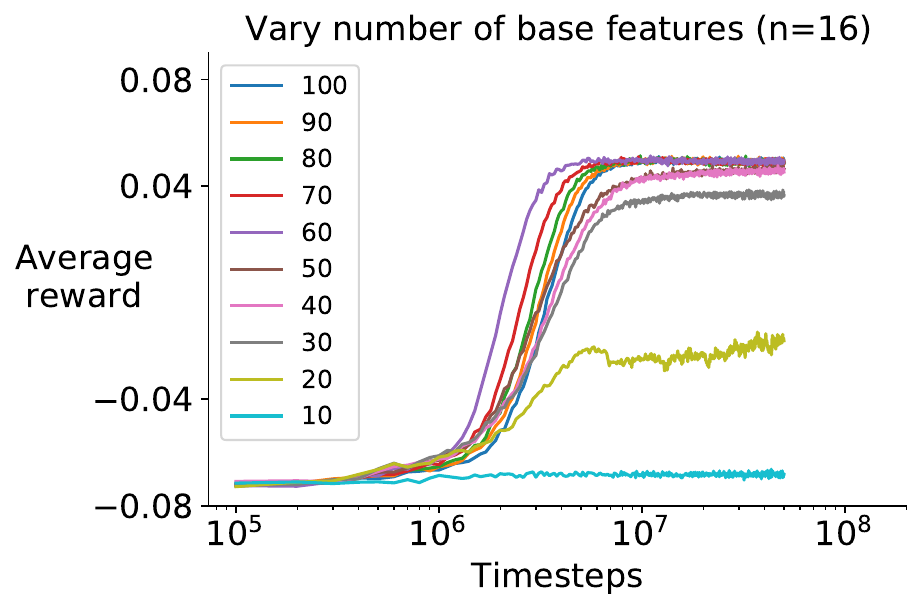}
    \caption{ On the left, we see a nibbler algorithm with a varying number of GVF questions applied to a single size coupled catch problem.  We see the performance collapses when not enough GVF questions are provisioned (the swap threshold was set to zero in these experiments).
    On the right, we see the sensitivity of the nibbler algorithm to the setting of the number of base features vs the number of base features of the problem (56).  For this environment, the performance remains robust to missing a few base features or having many extra base features.}
    \label{fig:vary_hypers}
\end{figure}

\subsection{Varying number of selected GVF questions on a fixed problem size}

We see the performance of the Nibbler algorithm as the scale of the answer resources are varied in Figure~\ref{fig:vary_hypers}(left).  When inadequate resources are available, performance collapses.  This suggests running with a higher threshold $\tau$ would be advantageous.

\subsection{Varying number of selected base features on a fixed problem size}

The performance consequence of varying the number of base features selected per GVF question is shown in Figure~\ref{fig:vary_hypers}(right).  Picking substantially fewer features ($\leq 30$) yields a substantial drop in  performance.  Including extra features ($\geq 70$) induces slightly slower learning, but similar final performance.
This completes our examination of the sensitivity of the Nibbler algorithm to the novel hyperparameter choices.   We did not study hyperparameter sensitivity to the more conventional deep RL hyperparameters (stepsize, momentum) as we expect they are more dependent on the base problem.

\end{document}